%% file: main.tex
\documentclass[10pt,twocolumn,letterpaper]{article}

\usepackage{cvpr}              % To produce the CAMERA-READY version
\input{preamble}
\definecolor{cvprblue}{rgb}{0.21,0.49,0.74}
\usepackage[pagebackref,breaklinks,colorlinks,allcolors=cvprblue]{hyperref}
\usepackage[accsupp]{axessibility}

\usepackage{multirow}

\def\paperID{20218} % *** Enter the Paper ID here
\def\confName{CVPR}
\def\confYear{2026}

\title{Disentangling to Re-couple: Resolving the Similarity-Controllability Paradox in Subject-Driven Text-to-Image Generation}

\author{Shuang Li\thanks{These authors contributed equally to this work.},\ Chao Deng\footnotemark[1],\ Hang Chen,\ Liqun Liu\thanks{Corresponding author.},\\ Zhenyu Hu,\ Te Cao,\ Mengge Xue,\ Yuan Chen,\ Peng Shu,\ Huan Yu,\ Jie Jiang\\
Tencent\\
{\tt\small \{shuangsali,kodideng,liqunliu\}@tencent.com}
}

\begin{document}
\renewcommand{\thefootnote}{\fnsymbol{footnote}}
\maketitle
\input{sec/0_abstract}
\input{sec/1_intro}
\input{sec/2_related}
\input{sec/3_method}
\input{sec/4_exp}
\input{sec/5_conclusion}
{
    \small
    \bibliographystyle{ieeenat_fullname}
    \bibliography{main}
}

% WARNING: do not forget to delete the supplementary pages from your submission 
\input{sec/X_suppl}

\end{document}

%% file: sec/0_abstract.tex
\begin{abstract}
Subject-Driven Text-to-Image (T2I) Generation aims to preserve a subject's identity while editing its context based on a text prompt. A core challenge in this task is the ``similarity-controllability paradox'', where enhancing textual control often degrades the subject's fidelity, and vice-versa. We argue this paradox stems from the ambiguous role of text prompts, which are often tasked with describing both the subject and the desired modifications, leading to conflicting signals for the model. To resolve this, we propose \textbf{DisCo}, a novel framework that first \textbf{Dis}entangles and then re-\textbf{Co}uples visual and textual information. First, our textual-visual decoupling module isolates the sources of information: subject identity is extracted exclusively from the reference image with the entity word of the subject, while the text prompt is simplified to contain only the modification command, where the subject refers to general pronouns, eliminating descriptive ambiguity. 
However, this strict separation can lead to unnatural compositions between the subject and its contexts. 
We address this by designing a dedicated reward signal and using reinforcement learning to seamlessly recouple the visually-defined subject and the textually-generated context. 
Our approach effectively resolves the paradox, enabling simultaneous high-fidelity subject preservation and precise textual control. Extensive experiments demonstrate that our method achieves state-of-the-art performance, producing highly realistic and coherent images.
\end{abstract}

%% file: sec/1_intro.tex
\section{Introduction}
\label{sec:intro}

Subject-Driven Text-to-Image (T2I) Generation~\cite{croitoru2023diffusion, ruiz2023dreambooth, gal2022image}, also referred to as text-to-image customization, is defined as the task that takes both \textit{textual prompts} and a \textit{given subject} as input, and generates an image that aligns with the textual prompts while preserving the subject's precise details. Compared to general text-to-image generation, this task enables user-customized image modifications and maintains the subject's fidelity and intellectual property (IP). It thus finds widespread applications in fields such as film production, advertising, and personalized recommendation systems.

\begin{figure}
    \centering
    \includegraphics[width=1\linewidth]{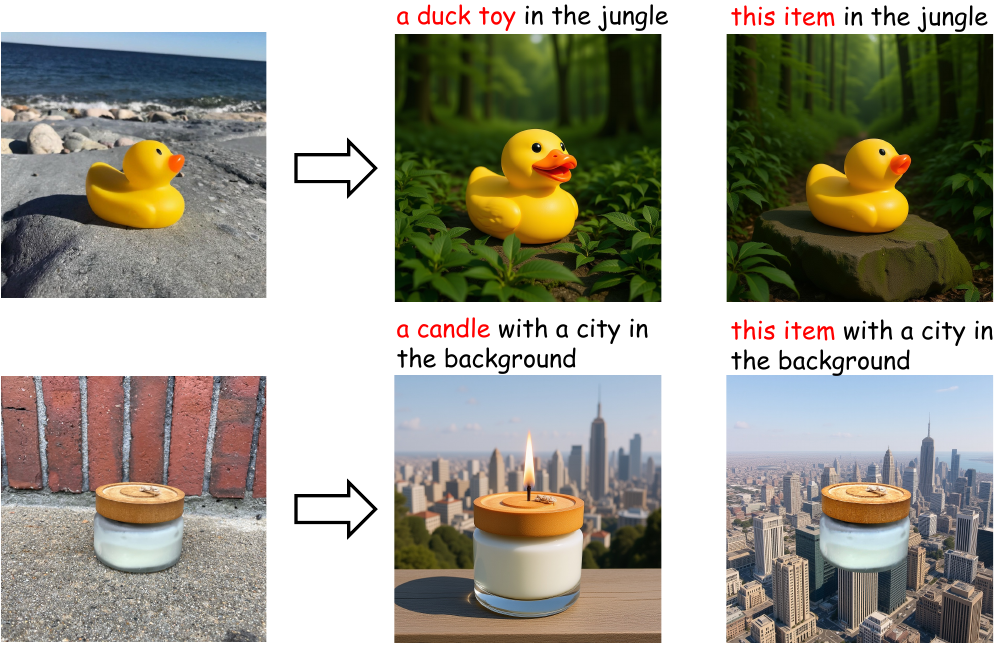}
    \caption{The results with different prompts of the same reference image generated by FLUX \textit{Kontext} [dev].}
    \label{fig:intro case}
\end{figure}

Unlike standard T2I generation, which primarily focuses on semantic adherence and image quality, the primary challenge of subject-driven T2I generation lies in navigating a fundamental trade-off: maintaining high similarity to the reference subject while achieving precise controllability from the text prompt. This tension gives rise to a dual-optimum paradox between \textit{similarity} (to the subject) and \textit{controllability} (by the prompt), where achieving optimal similarity to the given subject and optimal controllability via the text prompt cannot be optimized simultaneously~\cite{huang2024realcustom}.
% 可以删掉
To address this issue, several encoder-based methods like IP-Adapter~\cite{ye2023ip}, RealCustom++~\cite{mao2024realcustom++} and SSR-Encoder~\cite{zhang2024ssr} employ an encoder to extract the features of the subject facilitated by masks or entity words,  and inject them into the model with a specialized cross-attention mechanism. 
% However, these methods are primarily based on UNet-based diffusion models like SD1.5/XL~\cite{rombach2022high,podell2023sdxl}, which face limitations in overall generation quality and struggle to preserve fine-grained image details.
Recently, methods such as OminiControl~\cite{tan2024ominicontrol} and OminiGen~\cite{xiao2025omnigen} propose to integrate the subject image through a unified sequence with diffusion latents based on Diffusion Transformer (DiT)~\cite{peebles2023scalable}.
% the advent of Diffusion Transformer (DiT)~\cite{peebles2023scalable} has significantly scaled up the performance of text-to-image generation. 

% , , highlighting that DiT itself can serve as the image encoder for the subject.

While these work have achieved superior performance, the dual-optimum paradox between \textit{similarity} and \textit{controllability} remains a persistent bottleneck. 
We argue that the critical, yet overlooked, source of this paradox lies in the entanglement of roles between the text prompt and the reference image during inference.
Existing methods typically rely on prompts that describe both the subject and the desired modifications (e.g., \textit{``a duck toy in the jungle.''}). However, textual descriptions of a subject are inherently discrete and abstract, failing to capture the rich, continuous information of color, texture, and style present for the subject.
In this scenario, the model is also guided by the textual description of the subject in the prompt, rather than simply relying on its visual features of the reference image.
The entanglement creates a conflict: when the model prioritizes the prompt's description of the subject, it often compromises the visual fidelity of the reference image.

For example, ~\cref{fig:intro case} presents the results generated with different prompts of the same reference image.
% sourced from the DreamBench~\cite{ruiz2023dreambooth} dataset. 
% The results were generated by FLUX \textit{Kontext} [dev]\footnote{\textit{FLUX Kontext} [dev] is a 12 billion parameter rectified flow transformer capable of editing images based on text instructions.}, a powerful image editing model on the Artificial Analysis Image Arena Leaderboard\footnote{https://artificialanalysis.ai}.
The duck toy shown in the first row has a meticulously designed beak.
% is a classic test case, notable for its 
When using the prompt containing the phrase \textit{``a duck toy,''} the model's prior knowledge, acquired during training, overrides the reference image. 
This results in the generation of a generic duck toy, whose beak shape and wing details differ from the original subject.
However, when replacing \textit{``a duck toy''} with a non-descriptive generic pronoun (e.g., \textit{``this item''} or \textit{``it''}), the results maintain high similarity to the subject in the reference image. 
A similar behavior is observed with the candle sample in the second row. With the entity word \textit{``candle''}, the model draws upon its prior knowledge and defaults to generating a lit candle. 
% Although the resulting image incorporates features from the subject, this preconceived bias leads to significant alterations, particularly in the material texture and the addition of a wick.
This demonstrates that the model's failure in subject identity preservation may stem not from its capabilities, but can be caused by inaccurate or overly specific subject descriptions in the prompt that introduce conflicting priors.

We posit that for subject-driven T2I generation, \textit{subject identity should be derived exclusively from the reference image, while the text prompt should solely dictate the desired modifications}. Therefore, we propose a novel textual-visual decoupling module. In this module, the prompt is strategically simplified to remove any descriptive information about the subject, instead using a general pronoun (e.g., \textit{``this item''} or \textit{``it''}) to refer to the subject and focusing entirely on the modification instruction. Meanwhile, the subject in the reference image is identified with the corresponding entity word (e.g., \textit{``a duck toy''}). The subject identity is injected directly and completely from the visual features of the reference image. However, this strict decoupling introduces another challenge: the subject, defined visually, and the context, defined textually, lack the necessary interaction, leading to unnatural compositions.  
As illustrated in the second example of ~\cref{fig:intro case}, the original prompt generates a plausible scene where a candle is placed on an surface against a city landscape. However, once we decouple the textual and visual inputs, the model no longer receives contextual information about the subject from the prompt. Consequently, it struggles to determine how to compose ``this item'' within the scene described as ``a city in the background.'' While the subject fidelity is well maintained, the model defaults to the subject's relative size and position from the reference image, erroneously placing the candle in mid-air. This leads to a physically implausible composition.

% 这句话要放在后面吗
To overcome the compositional gap, we leverage Group Relative Policy Optimization (GRPO) to enhance the coupling of textual and visual features. The successful adaptation of GRPO to T2I models by \citep{xue2025dancegrpo, liu2025flow} has already demonstrated significant gains in generation quality.
However, applying reinforcement learning to the specific task of subject-driven T2I generation introduces unique challenges. While existing reward metrics like ImageReward \cite{xu2023imagereward}, CLIP-T \cite{radford2021learning}, and HPS \cite{wu2023human} are effective for general T2I, they primarily focus on overall image quality or text-image alignment. The subject-driven task requires a more comprehensive reward signal. Specifically, an effective reward model must be able to assess not only text alignment but also subject similarity, as well as the compositional harmony of the subject's interaction with the context.

To address this, we propose a novel strategy for training a discriminative reward model. We utilize a Vision Language Model (VLM) to automatically generate editing instructions. These instructions are designed to synthetically create negative examples by modifying the target image's subject ID features or its interaction with the context. By constructing preference pairs with original and edited images, we train a reward model that becomes highly adept at distinguishing these precise failure modes. This new reward model provides a more robust and accurate signal, guiding the GRPO process.

% While reinforcement learning (RL) has recently been adopted to align T2I models with human preferences and aesthetic quality~\cite{wallace2024diffusion,liu2025flow,zhang2024onlinevpo},
% we turn to 
% GRPO is a powerful policy optimization method in reinforcement learning, originally developed to enhance LLMs on complex reasoning tasks\cite{guo2025deepseek}.

% 
% In our framework, GRPO refines the model by optimizing for rewards that reflect generation quality and physical plausibility. This process enhances the coupling between the subject and context, guiding the model to generate plausible physical interactions and harmonious compositions, ensuring the final image is both coherent and realistic.

In summary, we propose \textbf{DisCo}, a novel \textbf{Dis}entangle and re-\textbf{Co}uple framework. DisCo first separates subject identity from textual control and then employs GRPO to re-couple them for natural and coherent subject-driven image generation. This resolves the similarity-controllability paradox, enabling the model to generate high-quality images that maintain subject fidelity while precisely adhering to the prompt.
Our main contributions are summarized as follows:
\begin{itemize}
    \item We identify the overloaded text prompt as a primary cause of the \textit{similarity-controllability paradox} in customized image generation. We propose a novel textual-visual decoupling module that isolates subject identity (sourced from the reference image) from contextual control (dictated by the prompt) to resolve this conflict.
    \item We train a dedicated reward model for subject-driven generation by automatically creating synthetic negative examples. This reward supervises the GRPO process, significantly improving the compositional harmony and re-coupling the subject with its context.
    \item Through extensive experiments, our method achieves state-of-the-art performance against existing approaches. 
    Furthermore, we conducted detailed ablation studies to verify the effectiveness of the visual-textual decoupling module, the dedicated reward model and GRPO.
\end{itemize}

%% file: sec/2_related.tex
\section{Related Work}
\label{sec:related work}

\subsection{Subject-driven Text-to-image Generation}
Early work~\cite{ruiz2023dreambooth, gal2022image, miao2024rpo} follow the pseudo-word paradigm by finetuning the pre-trained models for each subject individually. These methods require extensive resources for tuning individual instances which renders them less feasible for widespread application. 
Some methods~\cite{ye2023ip, wei2023elite, li2023blip, wang2024ms} have sought to employ an encoder for subjects without additional fine-tuning.
IP-Adapter~\cite{ye2023ip} encodes subjects into text-compatible image prompts for subject personalization. 
MS-diffusion~\cite{wang2024ms} introduces the layout-guided zero-shot image personalization with multiple subjects framework. 
RealCustom~\cite{huang2024realcustom, mao2024realcustom++} and SSR-Encoder~\cite{zhang2024ssr} selectively inject subject-related features from images to reduce background features.
While some works design dedicated attention mechanisms for subject-text interactions\cite{xiao2024fastcomposer,li2024photomaker,wang2024moa,patashnik2025nestedattention}, more recent study such as OminiControl~\cite{tan2024ominicontrol, tan2025ominicontrol2}, UniReal~\cite{chen2025unireal}, DreamO~\cite{mou2025dreamo} and others~\cite{xiao2025omnigen, wu2025less} explore the MM-Attention mechanism in DiTs~\cite{esser2024scaling} to incorporate image conditions with a unified sequence.
Despite these advances, these works can not solve the dual-optimum paradox between \textit{similarity} and \textit{controllability}.

\subsection{Reinforcement Learning with Diffusion Models}
Reinforcement learning (RL), first popularized in Large Language Models (LLMs)~\cite{guo2025deepseek,achiam2023gpt} using policy gradient methods like PPO~\cite{schulman2017proximal} or value-free GRPO~\cite{shao2024deepseekmath}, has recently become a key research area for diffusion models. 
Current methods can be classified into three main categories. The first is DPO-style optimization, which learns from preference data directly~\cite{wallace2024diffusion, zhang2025diffusion}. The second is direct reward optimization, where gradients from a reward signal are backpropagated through the model, such as ReFL~\cite{xu2023imagereward}. The third is policy gradient-based RL, which adapts classic RL algorithms like DDPO~\cite{black2023training} and DPOK~\cite{fan2023dpok}.  
GRPO is a recent breakthrough that enhances LLMs on complex reasoning tasks.
Flow-GRPO\cite{liu2025flow} and DanceGRPO\cite{xue2025dancegrpo} have successfully introduced GRPO to flow-based models, demonstrating significant improvements in image generation quality. In this paper, we adopt GRPO to re-couple visual and textual features. It enables the model to generate more realistic images with harmonious integration between the subject and background.

%% file: sec/3_method.tex
\section{Method}

\begin{figure}
    \centering
    \includegraphics[width=1\linewidth]{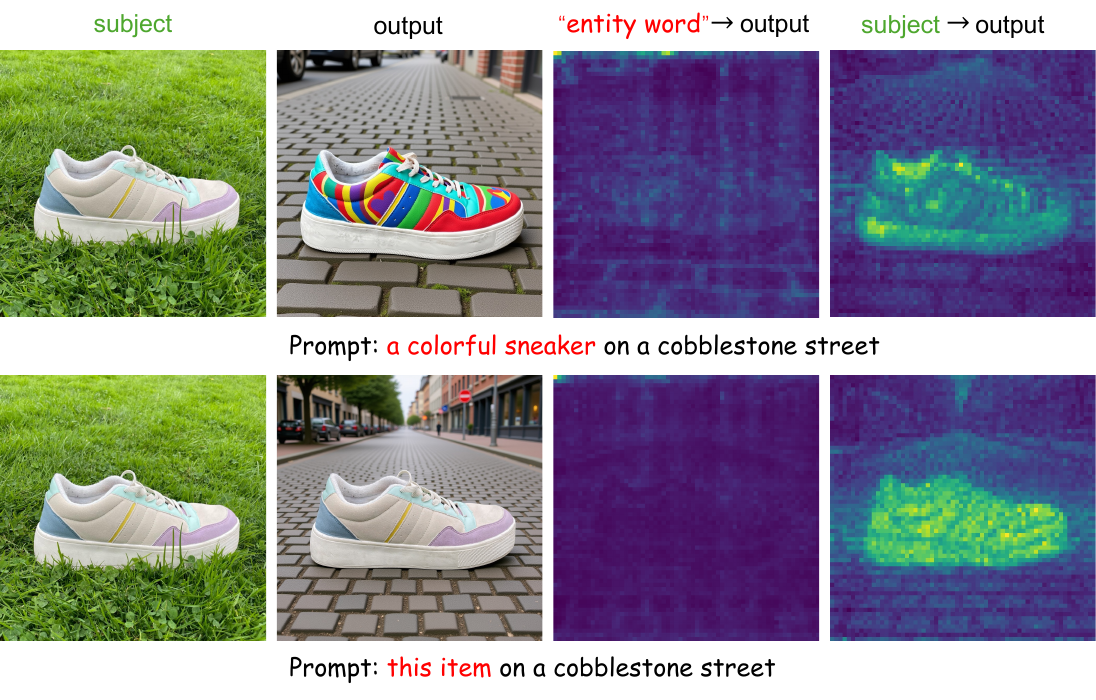}
    \caption{Visualization of the attention maps between entity word and subject to the generated image, respectively.}
    \label{fig:attention}
\end{figure}
\begin{figure*}
    \centering
    \includegraphics[width=0.9\linewidth]{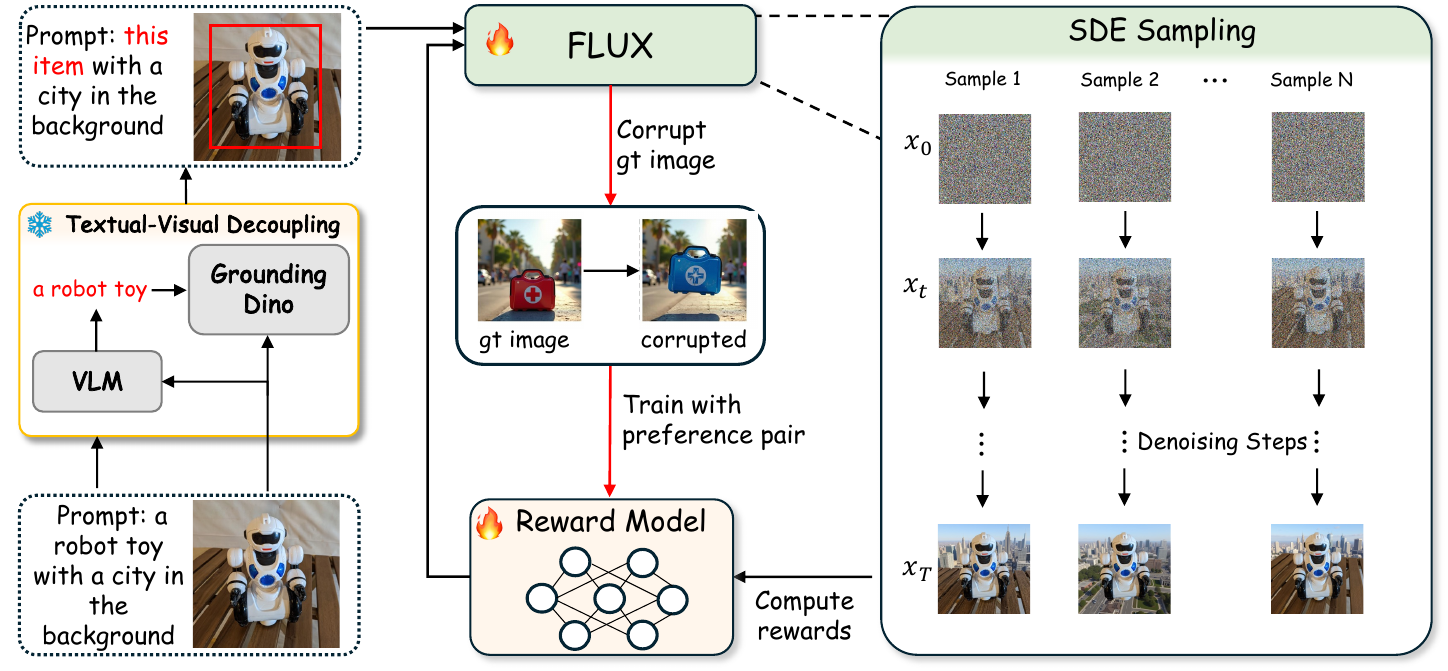}
    \caption{
    \textbf{The architecture of our proposed DisCo framework.} DisCo first decouples subject identity (from the image) from textual control (from the prompt). We generate corrupted images to construct preference pairs with the ground truth images to train the Reward Model. Subsequently, it employs GRPO to re-couple textual and visual features, generating a coherent image that preserves subject details while adhering to the prompt. }
    \label{fig:model}
\end{figure*}
\label{sec:method}
This section introduces the architecture of our proposed method, DisCo, which is illustrated in ~\cref{fig:model}.
% Section~\cref{subsec:pre} introduces the foundational work of DiT; In Section~\cref{subsec:decouple}, we introduce the textual-visual decoupling module designed to isolate subject identity from textual control; Section~\cref{subsec:grpo} explains how we employ the GRPO module to re-couple the visually-defined subject and the textually-generated background.

\subsection{Preliminary}
\label{subsec:pre}
The DiT model, employed in architectures like FLUX~\cite{flux2024} and Stable Diffusion 3~\cite{rombach2022high}, generates high-quality images through an iterative denoising process. In FLUX, at each denoising step, DiT blocks process a token sequence which consists of noisy image tokens $\mathbf{X} \in \mathbb{R}^{* \times d}$ and text condition tokens $\mathbf{C}_T \in \mathbb{R}^{* \times d}$, where $d$ is the embedding dimension and $*$ denotes the number of tokens.
The previous work~\cite{tan2024ominicontrol} extends the input sequence with image condition tokens for subject-driven T2I generation. For the reference image, it is first encoded into latent tokens $\mathbf{C}_I \in \mathbb{R}^{* \times d}$ using the VAE encoder. The extended token sequence becomes:
\begin{equation}
    \mathbf{S} = [\mathbf{X}; \mathbf{C}_T; \mathbf{C}_I]
\end{equation}

Following the setup of FLUX \textit{Kontext}, we employ a unified three-dimensional Rotary Position Embedding (3D RoPE)~\cite{su2024roformer} within each DiT block. Image tokens are encoded with the position $(t, h, w)$, where the virtual temporal index $t$ distinguishes the noise latent from the reference image, and $(h, w)$ denote the spatial coordinates. Text condition tokens are assigned a fixed position of $(0, 0, 0)$ to serve as a global condition.

The multi-modal attention then projects position-encoded tokens into query $Q$, key $K$, and value $V$ representations enabling attention computation across all tokens:
\begin{equation}
    \text{MMA}([\mathbf{X}; \mathbf{C}_T; \mathbf{C}_I]) = \text{softmax}\left(\frac{QK^\top}{\sqrt{d}}\right)V,
\end{equation}
where $[\mathbf{X}; \mathbf{C}_T; \mathbf{C}_I]$ denotes the concatenation of image, text and reference image tokens.

\subsection{Textual-visual Decoupling Module}
\label{subsec:decouple}

As mentioned in Introduction, the \textit{similarity-controllability paradox} is a critical bottleneck in subject-driven T2I generation. 
We argue that its primary cause is the textual descriptions of the subject entangle with modification instructions, activating the model's strong preconceived biases and compromising visual fidelity. 
To resolve this, we propose a textual-visual decoupling module guided by a core principle: \textit{subject identity should be derived exclusively from the reference image, while the text prompt should solely dictate the desired modifications}.

This module implements the principle through a systematic, two-part process, illustrated in \cref{fig:model}. Given an input prompt ($c_T$) and a reference image ($c_I$), the process begins as follows:
\paragraph{Prompt Simplification.} First, we must identify the source of the conflicting textual signal. We employ a powerful Vision-Language Model (Qwen2.5-VL 72B~\cite{bai2025qwen25vl}, to analyze the prompt in the context of the image and identify the subject. We term this the ``\textit{entity word}'' (e.g., ``a duck toy''). This entity word is precisely what triggers the model's internal, often generic, priors, as seen with the duck's beak and the candle's flame in ~\cref{fig:intro case}. Therefore, we strategically simplify the prompt by replacing the entity word and its related descriptions with a generic placeholder, such as ``\textit{this item}'' or ``\textit{it}''. The resulting prompt (e.g., ``\textit{this item in the jungle}'') forces the model to seek identity information exclusively from the visual modality, effectively silencing the conflicting textual guidance.

\paragraph{Visual Grounding.}
Simplifying the prompt, however, creates ambiguity: the model no longer knows which object in the reference image ``\textit{this item}'' refers to. To resolve this, we leverage GroundingDINO~\cite{liu2024grounding} to explicitly localize the subject in the reference image $c_i$. Crucially, we provide GroundingDINO with the original \textit{entity word} identified in the previous step. This allows for precise localization of the intended subject without altering the reference image itself. 
This process effectively bridges the generic pronoun in the decoupled prompt with the specific, high-fidelity visual features of the subject in the reference image, ensuring precise identity preservation.

\paragraph{Analysis of Attention Maps.}
To empirically validate how this module alters the interaction between latents, text and reference image, we visualize the cross-attention maps from DiT block in ~\cref{fig:attention}.  For each row, we compute the attention from the conditions (the entity word from the original prompt and the grounded subject from the image) to the generated image. The top row, representing the baseline with entangled signals, reveals that the entity word (``a colorful sneaker'') casts a strong attentional focus on the subject, confirming its undesirable influence. In contrast, after applying our decoupling module in the bottom row, the attention from the now-absent entity word is suppressed. Instead, the attention map originating from the grounded reference subject now precisely and strongly targets the corresponding area in the generated image. This provides direct evidence that our method successfully shifts the locus of control for identity from the overloaded text prompt to the high-fidelity visual reference, leading to a well-preserved subject.

\subsection{Textual-Visual Re-Coupling}
%% 简单介绍flow-grpo -> 展开reward modeling
\label{subsec:grpo}
Following the textual-visual decoupling, a significant challenge remains: \textit{the compositional gap}. The model, having learned to treat the subject's identity and the background context as separate entities, may struggle to merge them into a single, coherent image. 
To address this gap, we introduce Group Relative Policy Optimization (GRPO) as a second stage and design a reward model that co-evolves with the image generation model.

While existing reward models such as ImageReward, HPSv2, and CLIP-based scorers primarily focus on assessing overall image quality, subject-driven T2I generation requires a richer notion of reward. In particular, the model must simultaneously evaluate (i) \textit{subject similariity} between the generated image and the reference image, and (ii) \textit{compositional integration} between the subject and context. To this end, we train a specialized reward model based on Qwen3-VL-30B, tailored for this task.
% To further enhance the reward model’s ability to detect subject-level inconsistencies and compositional failures, 
We construct the preference data using synthetically generated negative examples.
We utilize a VLM to produce editing instructions that intentionally modify either the subject identity features or the interaction between the subject and the surrounding context. Applying these instructions yields edited images that exhibit controlled deviations from the desired subject appearance or compositional fidelity.
Given a reference image, the VLM produces an editing instruction, which is applied to obtain an edited image. For each original-edited pair, The model is trained to assign higher preference to the original image, which preserves the target subject identity and composition. Additional implementation details are provided in the Appendix.
% the reward model receives both images:
% \begin{equation}
% p_\phi(I \succ \tilde{I} \mid I_{\mathrm{ref}}, x, e)
% = R_\phi(I, \tilde{I}, I_{\mathrm{ref}}, x, e).
% \end{equation}

% Given a dataset of preference pairs $\mathcal{D} = {(I^+, I^-)}$, we optimize ($\phi$) via the negative log-likelihood:
% \begin{equation}
% \mathcal{L}*{\mathrm{RM}}(\phi)
% = - \mathbb{E}*{(I^+, I^-)\sim\mathcal{D}}
% \left[
% \log p_\phi(I^+ \succ I^- \mid I_{\mathrm{ref}}, x)
% \right].
% \end{equation}

% In this formulation:
% \begin{itemize}
%     \item The \textbf{state} $s_t$ at timestep $t$ is the noisy image latent $x_t^i$.
%     \item The \textbf{action} $a_t$ is the model's prediction of the subsequent, less noisy state $x_{t-1}^i$.
%     \item The \textbf{policy} $\pi_{\theta}$ is the flow-based model $p_{\theta}(x_{t-1}^i | x_t^i, c)$, which guides the reverse-time trajectory.
%     \item The \textbf{reward} $R(x_0^i, c)$ is a score assigned to the generated image $x_0^i$.
% \end{itemize}

In GRPO stage, for the given pair $c$ of a prompt $c_T$ and a reference image $c_I$, we first sample a group of $G$ images $\{x_0^i\}_{i=1}^G$, and construct pairwise comparisons. Each image pair, together with the reference image, is fed into the reward model, which selects the preferred image. We sum the log-probabilities associated with the tokens that support its selection and this aggregated value serves as the final reward for that image. 
 
The advantage $\hat{A}_t^i$ for the $i$-th image within each group is then calculated by normalizing its final reward against the mean and standard deviation of the entire group's rewards:
\begin{equation}
    \label{eq:advantage}
    \hat{A}_t^i = \frac{R(x_0^i, c) - \text{mean}(\{R(x_0^i, c)\}_{i=1}^G)}{\text{std}(\{R(x_0^i, c)\}_{i=1}^G)}.
\end{equation}
With the advantage defined, GRPO optimizes the policy model by maximizing the following objective function:
\begin{equation}
    \label{eq:grpo_objective}
    \mathcal{J}_{\text{GRPO}}(\theta) = \mathbb{E}_{c \sim C, \{x^i\}_{i=1}^G \sim \pi_{\theta_{\text{old}}}(\cdot|c)} f(r, \hat{A}, \theta, \varepsilon, \beta),
\end{equation}
where
\begin{align}
    \resizebox{1\linewidth}{!}{$
    \begin{aligned}
    f(\cdot) 
    &= \frac{1}{G} \sum_{i=1}^G \frac{1}{T} \sum_{t=0}^{T-1} \biggl[ 
       \min\Bigl( r_t^i(\theta)\hat{A}_t^i, \text{clip}\bigl(r_t^i(\theta), 1-\varepsilon, 1+\varepsilon\bigr)\hat{A}_t^i \Bigr) \\
    &\quad - \beta \mathbb{D}_{\text{KL}}\bigl(\pi_{\theta} \| \pi_{\text{ref}}\bigr) \biggr],
    \end{aligned}
    $} 
\label{eq:grpo_f}
\end{align}

\begin{equation}
r_t^i(\theta) = \frac{p_{\theta}(x_{t-1}^i | x_t^i, c)}{p_{\theta_{\text{old}}}(x_{t-1}^i | x_t^i, c)}\label{eq:importance_ratio}.
\end{equation}

%% file: sec/4_exp.tex
\section{Experiments}
\label{sec:experiments}

\subsection{Experimental Setups}
\label{subsec:setup}
\paragraph{Implementation Details.} We adopt FLUX~\cite{flux2024} as the base model and train on Subjects200K dataset proposed by \citet{tan2024ominicontrol}.
For the training process, we employ the AdamW~\cite{loshchilov2017decoupled} optimizer with a learning rate of 1e-5 and train on 8 NVIDIA H20 80G GPUs. The batch size is set to 2 and the gradient accumulation step is set to 12. In the settings of GRPO, the sampling timestep is set to 16 and the numbers of generation image per prompt is set to 12. The noise level $\epsilon$ is set to 0.3. We adopt Qwen3-VL-30B~\cite{bai2025qwen3vl} as the reward model, trained with 25k preference pairs constructed from original and edited images generated by FLUX. We will release both the dataset used for training and the code of our method. Additional training details of the reward model can be found in the Appendix.

\paragraph{Baselines.} We compare our method with following methods: (1) SDXL-based methods: IP-Adapter~\cite{ye2023ip}, Emu2~\cite{Emu2}, SSR-Encoder~\cite{zhang2024ssr}, MS-Diffusion~\cite{wang2024ms} and RealCustom++~\cite{mao2024realcustom++}; (2) FLUX-based methods: OminiControl~\cite{tan2024ominicontrol}, ACE++~\cite{mao2025ace++},  UNO~\cite{wu2025less}, DreamO~\cite{mou2025dreamo}, and FLUX \textit{Kontext} [dev]~\cite{labs2025flux1kontextflowmatching}. For a fair comparison, we evaluate all baseline models by running their official open-source code on the benchmark, using the default hyper-parameter settings as specified in the respective papers.

\begin{figure*}[t]
    \centering
    \includegraphics[width=0.98\linewidth]{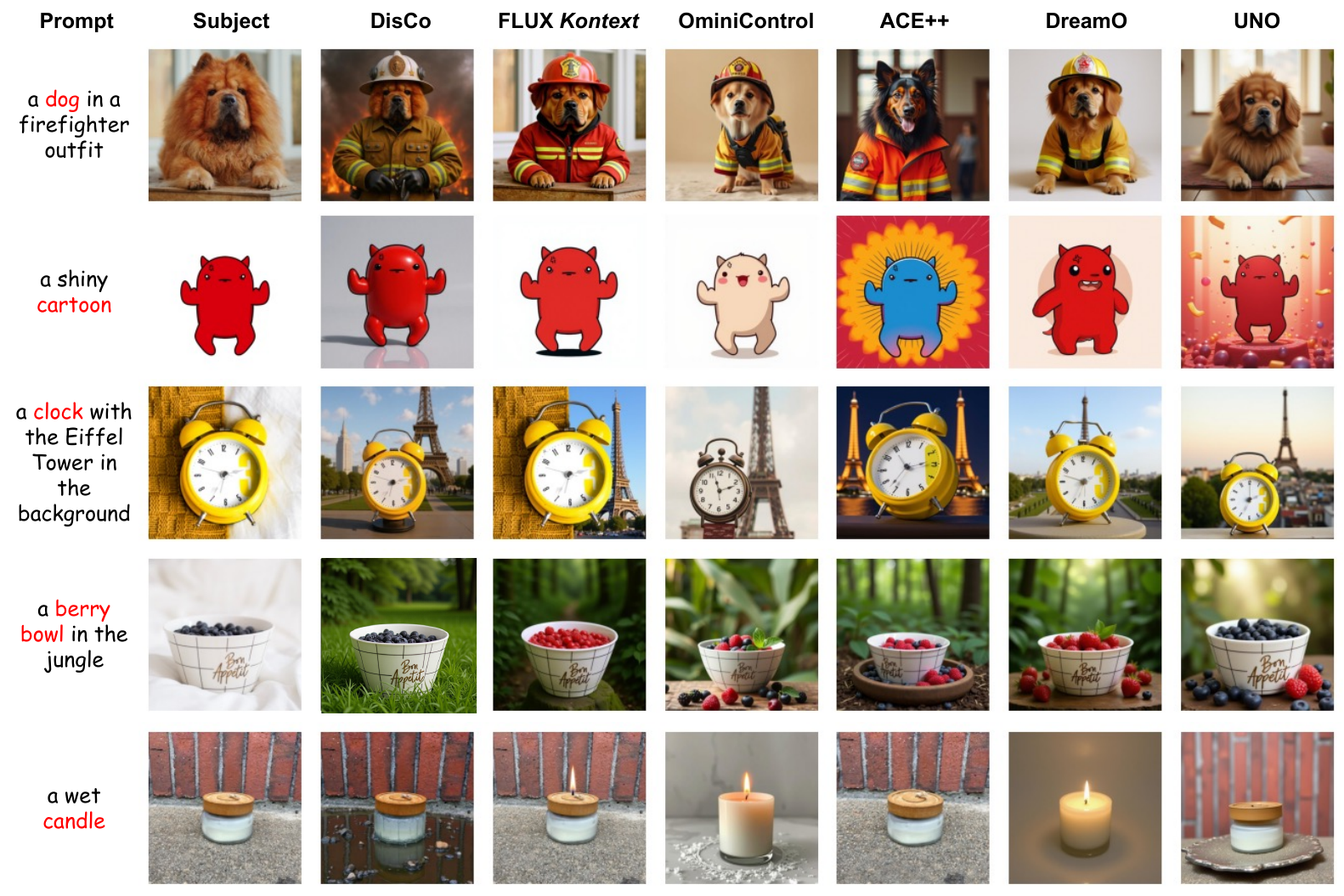}
    \caption{Qualitative comparison with baseline methods on DreamBench.}
    \label{fig:qualitative}
\end{figure*}
\paragraph{Evaluation Metrics.}
We conduct experiments on DreamBench~\cite{ruiz2023dreambooth}. The benchmark includes 30 subjects with 25 editing prompts each, leading to 750 evaluation cases.
\textbf{\textit{Subject similarity}} is measured by CLIP-I and DINO~\cite{caron2021emerging} scores. Following RealCustom++, we use SAM~\cite{kirillov2023segment} to segment the main subject from both the reference and generated images before computing the similarity scores. \textbf{\textit{Text controllability}} is evaluated by the text-image embedding similarity with CLIP~\cite{radford2021learning}. For CLIP-based metrics, we use two CLIP model variants: ViT-B/32 and ViT-L/14, denoted as CLIP-B-* and CLIP-L-*, respectively. We also adopt ImageReward~\cite{xu2023imagereward} as an overall metric for \textbf{\textit{Image quality}}.

\subsection{Main Results}
\begin{table*}[!htbp]
    \caption{Quantitative comparison of DisCo and baselines on DreamBench. Bold and underline represent the highest and second-highest metrics, respectively.}
    \label{tab:comparison}
    \centering
    \resizebox{1\textwidth}{!}{
    % I have removed the 'BaseModel' column and adjusted the column count and cmidrule indices.
    \begin{tabular}{l ccc cc c}
    \toprule
    \multirow{2}{*}{Method} &
    \multicolumn{3}{c}{Subject Similarity} &
    \multicolumn{2}{c}{Text Controllability} &
    \multicolumn{1}{c}{Image Quality} \\
    \cmidrule(lr){2-4}\cmidrule(lr){5-6}\cmidrule(lr){7-7}
    & CLIP-B-I$\uparrow$ & CLIP-L-I$\uparrow$ & DINO-I$\uparrow$
      & CLIP-B-T$\uparrow$ & CLIP-L-T$\uparrow$
      & ImageReward$\uparrow$ \\
    \midrule
    IP-Adapter    & 0.839 & 0.823 & 0.684 & 0.311 & 0.233 & 0.463  \\
    Emu2          & 0.833 & 0.813 & 0.719 & 0.291 & 0.235 & -0.148  \\
    SSR-Encoder   & 0.832 & 0.804 & 0.730 & 0.302 & 0.251 & 0.139 \\
    MS-Diffusion  & 0.814 & 0.795 & 0.727 & 0.312 & 0.259 & 1.098 \\
    RealCustom++  & 0.801 & 0.772 & 0.749 & 0.315 & 0.258 & 1.251 \\
    OminiControl & 0.857 & 0.867 & 0.713 & 0.320 & 0.263 & 1.254 \\
    ACE++        & 0.876 & 0.881 & 0.734 & 0.321 & 0.267 & 1.150 \\
    DreamO       & 0.899 & 0.901 & 0.813 & \underline{0.322} & 0.267 & 1.186 \\
    UNO          & 0.899 & 0.907 & 0.827 & 0.311 & 0.255 & 0.854 \\
    FLUX.1 Kontext [dev] & \underline{0.910} & \underline{0.911} & \underline{0.839} & 0.321 & \underline{0.268} & \underline{1.276} \\
    % \textbf{DisCo (ours)} & \textbf{0.914} & \textbf{0.920} & \textbf{0.873} & \textbf{0.326} & \textbf{0.271} & \textbf{1.404} \\
    % 1105-ckpt25  & 0.930 & 0.934 & 0.915 & 0.319 & 0.265 & 1.320 \\
    % 1105-ckpt70  & 0.930 & 0.937 & 0.917 & 0.318 & 0.263 & 1.147 \\
    \textbf{DisCo (ours)}  & \textbf{0.928} & \textbf{0.937} & \textbf{0.903} & \textbf{0.329} & \textbf{0.273} & \textbf{1.339} \\
    % 1110-ckpt38  & 0.918 & 0.924 & 0.894 & 0.321 & 0.265 & 1.189 \\ 
    % 1111-ckpt12  & 0.921 & 0.926 & 0.892 & 0.320 & 0.265 & 1.191 \\
    % 1111-ckpt25  & 0.920 & 0.929 & 0.896 & 0.319 & 0.263 & 1.150 \\
    % 1111-ckpt39  & 0.920 & 0.929 & 0.897 & 0.317 & 0.259 & 1.088 \\
    \bottomrule
    \end{tabular}
    }
\end{table*}

\paragraph{Quantitative Results.} As presented in \cref{tab:comparison}, our proposed method, DisCo, demonstrates superior performance over both SDXL-based and FLUX-based baselines across all evaluation metrics.
For subject similarity, DisCo achieves state-of-the-art results. It obtains the highest scores for all of CLIP-B-I (0.928), CLIP-L-I (0.937) and DINO-I (0.903). These results empirically validate our approach of decoupling textual and visual conditions, which encourages the model to source subject-related features primarily from the reference image.
Notably, DisCo also excels in text controllability, a metric that often presents a trade-off with subject similarity. This tension is evident in prior work; for instance, the results for SSR-Encoder vs. MS-Diffusion and DreamO vs. UNO show that improving one metric can lead to the degradation of the other. In contrast, our method effectively mitigates this trade-off, achieving the highest scores for both subject similarity and text controllability simultaneously.
Finally, DisCo maintains plausible image quality while resolving the core simliarity-controllability paradox. It achieves an IR score 1.339, outperforming the next-best competitor, FLUX \textit{Kontext}.

\paragraph{Qualitative Results.} \Cref{fig:qualitative} provides a qualitative comparison of DisCo against FLUX-based methods. 
Flux \textit{Kontext} excels other baselines in terms of subject similarity, but this model has drawbacks. 
It often copies the reference subject into a new scene with minimal integration, leading to unnatural results as seen in the dog and clock cases.
Moreover, because its prompt contains entangled descriptive information, it can also exhibit poor subject fidelity in certain scenarios such as the berry bowl.
OminiControl and ACE++ frequently fail to preserve the subject's identity, altering key attributes like the dog's breed and the cartoon monster's color.
DreamO and UNO show improved subject similarity but suffer from limited text controllability. 
For instance, UNO completely ignores the prompt ``in a firefighter outfit'' for the dog example. 
Among all methods, DisCo achieves a superior balance of subject consistency, text controllability, and overall image quality.

\begin{figure}[htbp]
    \centering
    \includegraphics[width=1\linewidth]{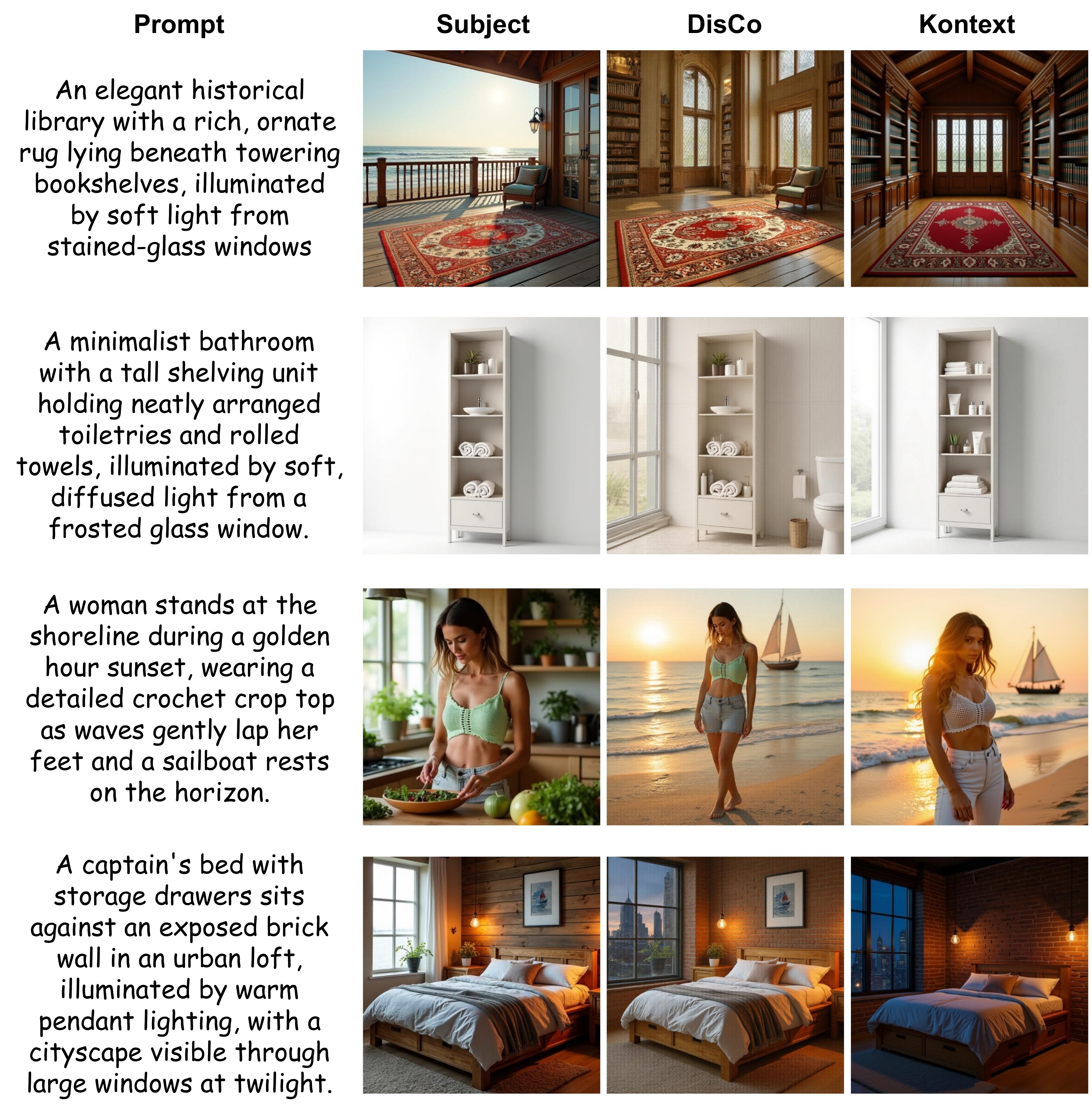}
    \caption{Qualitative results on complex scenarios.}
    \label{fig:complex}
\end{figure}

\begin{figure}[htbp]
    \centering
    \includegraphics[width=1\linewidth]{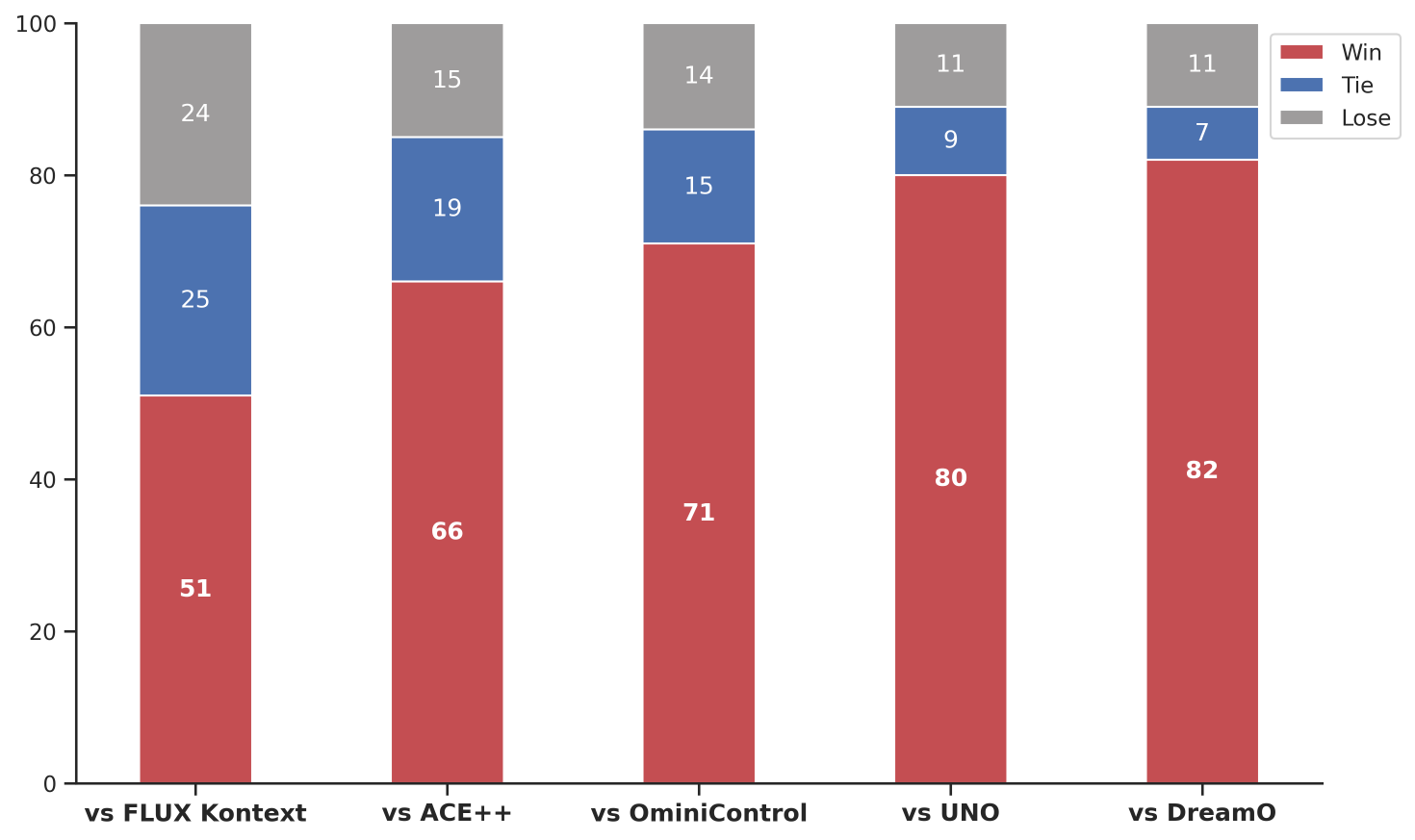}
    \caption{Pairwise human evaluation on 100 cases.}
    \label{fig:user_study}
\end{figure}

Furthermore, DisCo's advantages become even more pronounced in complex scenarios. When compared against the best baseline, FLUX \textit{Kontext}, \cref{fig:complex} illustrates that our method accurately identifies and preserves the subject's visual features while integrating them into the new context.
\paragraph{User Study.} 
To validate the effectiveness of our method, we conducted a large-scale human evaluation, comparing DisCo against five leading methods via a pairwise, side-by-side study on 100 samples. 
Participants were asked to choose the better image or declare a tie based on overall quality, subject similarity and text controllability. 
As shown in \cref{fig:user_study}, DisCo consistently outperforms all baselines. It achieves overwhelming win rates against UNO (80\%) and DreamO (82\%) and maintains a clear advantage over OminiControl (71\%) and ACE++ (66\%). Notably, even against FLUX \textit{Kontext}, DisCo is preferred in the majority of cases (51\% win rate; 24\% lose rate). This human-centric evaluation confirms that DisCo produces significantly more appealing and accurate results than existing  models.

\subsection{Ablation Study}
\label{subsec:ablation}
\begin{figure*}[htbp]
    \centering
    \includegraphics[width=0.8\linewidth]{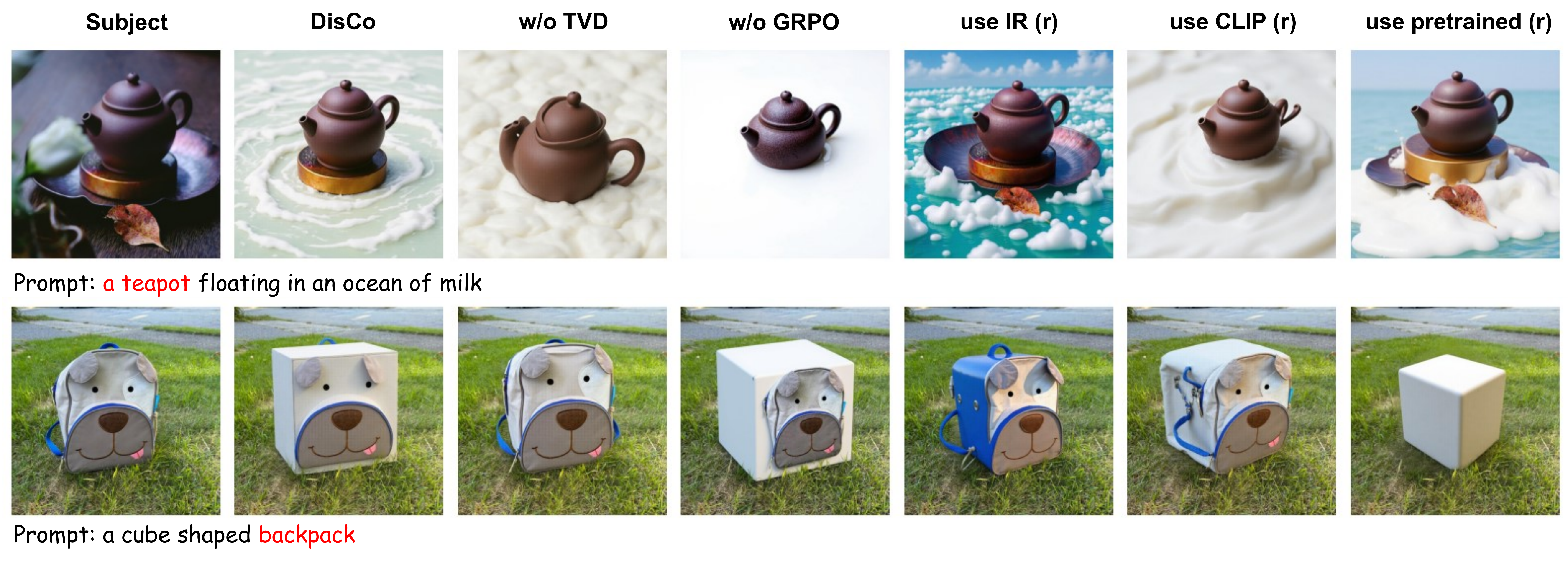}
    \caption{Comparison of image generation results by using different model settings.}
    \label{fig:alation_case}
\end{figure*}
As shown in \cref{tab:ablation}, we conduct a comprehensive ablation study to validate the effectiveness of the key components in DisCo. \Cref{fig:alation_case} illustrates several cases from results.
% \paragraph{Effect of Textual-Visual Decoupling Module.}  
We first evaluate the effect of textual-visual decoupling module by removing it (``w/o TVD''). In this setting, the model is trained on the original text prompt and subject. The results show a noticeable drop in both CLIP-I and CLIP-T scores compared to our full model. In \cref{fig:alation_case}, the results show that without decoupling, the model struggles to simultaneously preserve the subject's identity and follow the prompt. 
% \paragraph{Effect of the GRPO Stage.} 
Then we analyze the contribution of the GRPO stage by removing it and only using the output of the TVD module (``w/o GRPO'') to inference. This configuration yields the second-highest subject fidelity score (CLIP-I: 0.922) and the lowest prompt adherence score (CLIP-T: 0.319), as the model strictly sources the subject's identity from the reference image without contextual control. 
Moreover, as shown in \cref{tab:ablation} and \cref{fig:alation_case}, this comes at a significant cost to image quality, with IR (1.189) score dropping sharply. This demonstrates that GRPO is essential for re-coupling the decoupled visual and textual features, thereby enhancing the overall quality and realism of the generated images.
% \paragraph{Choice of Reward Model.} 

We investigate the impact of different reward models within the GRPO stage by comparing our approach with several alternatives. As shown in \cref{tab:ablation}, we first evaluate reward models adotped in text-to-image generation: CLIP (ViT-B/32) and ImageReward (IR), denoted as w/ CLIP (r) and w/ IR (r). Using CLIP as the reward model degrades overall performance, which we attribute to its inability to assess fine-grained qualities like compositional harmony and physical plausibility, both critical for our task. While ImageReward yields a slight improvement in image quality, it comes at the cost of a significant drop in subject similarity. This result highlights the necessity of incorporating reference images into reward modeling for subject-driven generation. We then evaluate using a pretrained VLM directly as the reward model (w/ pretrained (r)). Although this improves subject similarity slightly compared to CLIP and IR, it leads to worse text adherence and image quality, demonstrating the difficulty in calibrating a general-purpose pretrained model to the complicated preference criteria in subject-driven generation.
\begin{table}[htbp]
    \centering
    \caption{Ablation study of DisCo on DreamBench. TVD represents Textual-visual Decoupling Module.}
    \label{tab:ablation}
    % \resizebox{1\linewidth}{!}{ 
    \begin{tabular}{lcccc}
    \toprule
    Method &  CLIP-I$\uparrow$ & CLIP-T$\uparrow$ & IR$\uparrow$ \\
    \midrule
    w/o TVD & 0.915 & 0.319 & 1.237 \\
    w/o GRPO & \underline{0.922} & 0.319 & 1.189 \\
    use CLIP (r)  & 0.898 &0.319 &1.163 \\
    use IR (r) & 0.914 & \underline{0.326} & \textbf{1.404} \\
    use pretrained (r) & 0.918 & 0.321 & 1.189 \\ 
    \midrule
    \textbf{DisCo (Ours)} & \textbf{0.928} & \textbf{0.329} & \underline{1.339} \\
    \bottomrule
    \end{tabular}
    % }
\end{table}

%% file: sec/5_conclusion.tex
\section{Conclusion}
\label{sec:conclusion}
In this paper, we addressed the fundamental similarity-controllability paradox that hinders progress in subject-driven text-to-image generation. We identified the root cause of this issue as the entanglement of roles within overloaded text prompts, where descriptions of the subject conflict with the visual information from the reference image. 
To resolve this, we proposed \textbf{DisCo}, a novel \textbf{Dis}entangle and re-\textbf{Co}uple framework. Our approach first employs a textual-visual decoupling module to isolate subject identity, sourcing it exclusively from the reference image, while the simplified text prompt dictates only the desired modifications. Subsequently, we introduce Group Relative Policy Optimization (GRPO) to intelligently re-couple the subject and background, overcoming the compositional gap.
% and ensuring the generation of physically plausible and harmoniously blended images.
Extensive experiments demonstrate that DisCo achieves state-of-the-art performance, significantly outperforming existing methods in subject similarity while precisely adhering to prompt.
% over existing methods.

%% file: sec/X_suppl.tex
\clearpage
\setcounter{page}{1}
\maketitlesupplementary

\section{More Baselines and Benchmark Results}

In \cref{tab:dreambench}, we compared againt DiptychPrompting\cite{shin2025diptych} and RPO\cite{miao2024rpo}. While RPO achieves high CLIP-T scores, it requires extensive per-instance tuning, making it significantly less feasible for large-scale applications than our approach. Regarding IC-LoRA, its implementation requires task-specific comfyui workflows; we aim to include these results in the final version.
\Cref{tab:dreambench_plus} shows results on DreamBench++. DisCo still outperforms major baselines in subject consistency and prompt adherence across a wider range of subjects.

\begin{table}[!htbp]
    \caption{Quantitative results on DreamBench.}
    \vspace{-4mm}
    \label{tab:dreambench}
    \centering
    \resizebox{1\linewidth}{!}{
    \begin{tabular}{lcccc}
\toprule
Method & CLIP-B-I $\uparrow$ & DINO-I $\uparrow$ & CLIP-B-T $\uparrow$ & ImageReward $\uparrow$ \\
\midrule
DreamO & 0.899 & 0.813 & 0.322 & 1.186 \\
DreamO-decouple & 0.871 & 0.771 & 0.312 & 0.661 \\
UNO & 0.899 & 0.827 & 0.311 & 0.854 \\
UNO-decouple & 0.891 & 0.828 & 0.303 & 0.359 \\
Diptych Prompting & 0.864 & 0.767 & 0.319 & 1.136 \\
RPO & 0.852 & 0.725 & \textbf{0.338} & 1.164 \\
DisCo (ours) & \textbf{0.928} & \textbf{0.903} & \underline{0.329} & \textbf{1.339} \\
\bottomrule
    \end{tabular}
    }
\end{table}

\vspace{-3mm}
\begin{table}[!htbp]
    \caption{Quantitative results on DreamBench++.}
    \vspace{-4mm}
    \label{tab:dreambench_plus}
    \centering
    \resizebox{1\linewidth}{!}{
    \begin{tabular}{lcccc}
\toprule
Method & CLIP-B-I $\uparrow$ & DINO-I $\uparrow$ & CLIP-B-T $\uparrow$ & ImageReward $\uparrow$ \\
\midrule
OminiControl & 0.757 & 0.538 & 0.335 & \underline{1.173} \\
ACE++ & 0.787 & 0.552 & 0.324 & 0.762 \\
DreamO & \underline{0.795} & \underline{0.589} & \underline{0.336} & 1.044 \\
UNO & 0.782 & 0.549 & 0.321 & 0.693 \\
FLUX.1 Kontext [dev] & 0.790 & 0.572 & 0.334 & 1.118 \\
DisCo (ours) & \textbf{0.801} & \textbf{0.610} & \textbf{0.339} & \textbf{1.291} \\
\bottomrule
    \end{tabular}
    }
\end{table}

\section{Qualitative Results with General Text-to-image Models}

\begin{figure}[htbp]
    \centering
    \includegraphics[width=1\linewidth]{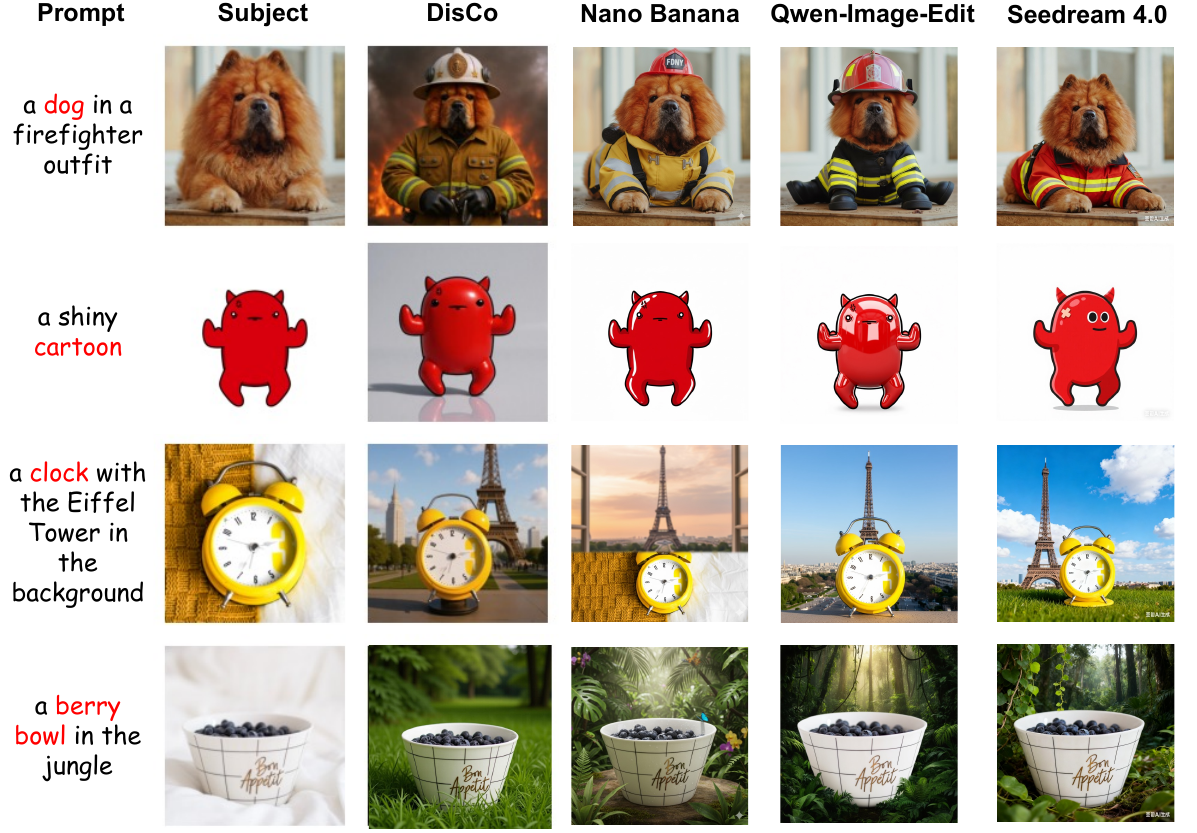}
    \caption{Qualitative results with leading general image editing models.}
    \label{fig:qualitative_general}
\end{figure}

\Cref{fig:qualitative_general} provides a qualitative comparison of DisCo against several leading general image editing models: Nano Banana, Qwen-Image-Edit\cite{wu2025qwenimage}, and Seedream 4.0\cite{seedream2025seedream4}. Our method achieves a superior balance of subject similarity and text controllability, showing performance that is comparable to or even surpasses these state-of-the-art methods.

However, other baselines exhibit specific drawbacks. Nano Banana occasionally compromises image authenticity; for instance, in the \texttt{clock} example, it spliced half of the reference image with the Eiffel Tower, generating a fabricated photograph. Qwen-Image-Edit struggles with maintaining correct proportions, as seen in the \texttt{dog} example where the scale appears disproportionate. Furthermore, Seedream 4.0 demonstrates relatively poor subject fidelity.
Among all methods, DisCo successfully mitigates these specific weaknesses, achieving the most balanced performance across subject consistency, seamless integration, and overall image quality.

\begin{figure}[htbp]
    \centering
    \includegraphics[width=1\linewidth]{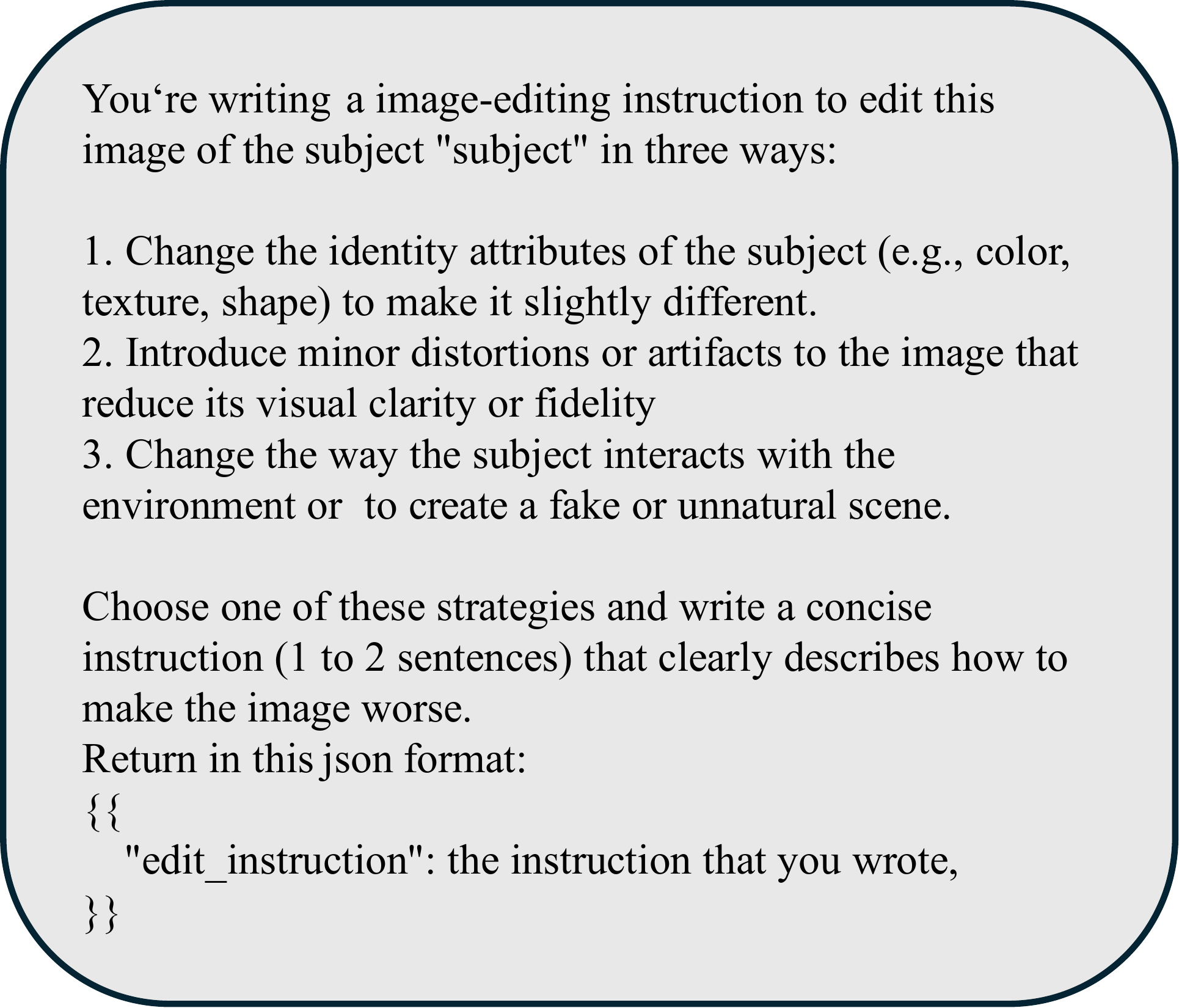}
    \caption{The prompt used to generate editing instructions.}
    \label{fig:prompt_corrupt}
\end{figure}

\section{Details of Reward Modeling}
% 评估主体驱动生成图像的结果需要综合的能力，因此我们选择使用VLM作为奖励模型。直接使用单点评估会因为对prompt过于敏感而导致评估噪声，因此我们在GRPO框架内构建了一个pairwise的奖励建模，可以形式化地表达为：R_i = sum_{j=1\cdots G, j\neq i}{P_{\phi}(x_0^i > x_0^j)}
Evaluating subject-driven text-to-image tasks requires a comprehensive understanding of both images and text instructions. In this work, we employ a Vision Language Model (VLM) for reward modeling, capitalizing on its multimodal understanding capabilities. While a straightforward approach would be pointwise evaluation (i.e., scoring a single image), such methods are often sensitive to minor prompt variations and can produce noisy reward signals. Our reward modeling, therefore, builds on a pairwise comparison framework within the Group Relative Policy Optimization (GRPO) stage.

The core of this framework is the computation of the preference probability $P_{\phi}(x^i \succ x^j \mid c_I, c_T)$, which indicates that the generated image $x^i$ is preferred over $x^j$. $c_I$ and $c_T$ are the reference image and text instruction, respectively, and $\phi$ is a preference predictor parameterized by a VLM. We compute this probability by presenting both images, $x^i$ and $x^j$, to a trained VLM $\phi$ and prompting it to select the superior one given $c_I$ and $c_T$. The normalized probability of the output token corresponding to the choice ``$x^i$ is better" is then used as the value for $P_{\phi}(x^i \succ x^j \mid c_I, c_T)$.

With this mechanism, the reward for a sample $x_0^i$ within a group of size $G$ during GRPO training is defined as the sum of its win probabilities against all other samples in the group:
\begin{equation}
R_i = \sum_{\substack{1 \le j \le G \\ j \ne i}} P_{\phi}(x_0^i \succ x_0^j \mid c_I, c_T)    
\end{equation}

To get more robust preference predictions, we build a dataset of preference pairs and train the VLM $\phi$. Instead of relying on manual annotation, we generate a synthetic dataset using the image generation model itself. 
For a given ``positive" example $(c_I, c_T, x_0)$ from the dataset, a VLM first crafts a ``negative" editing instruction $\tilde{c}_T$. This instruction intentionally introduces flaws, such as altering the subject's key attributes, creating inconsistencies with the original prompt $c_T$, or producing unnatural compositions. Subsequently, we generate a negative sample $\tilde{x}_0$ by applying the editing instruction $\tilde{c}_T$ to the original image $x_0$ using the image generation model. This process yields a preference pair $(x_0 \succ \tilde{x}_0)$, where the original image is preferred over the corrupted one.

The VLM $\phi$ is then trained by minimizing the negative log-likelihood of these preferences:
\begin{equation}
\mathcal{L}_{\phi} = \mathbb{E}_{(x_0,\tilde{x}_0,c_I,c_T)\sim\mathcal{D}}\left[-\log P_{\phi}(x_0 \succ \tilde{x}_0 \mid c_I, c_T)\right],
\end{equation}
where $\mathcal{D}$ is our synthetically generated dataset of preference tuples.

\begin{figure}[thpb]
    \centering
    \includegraphics[width=1\linewidth]{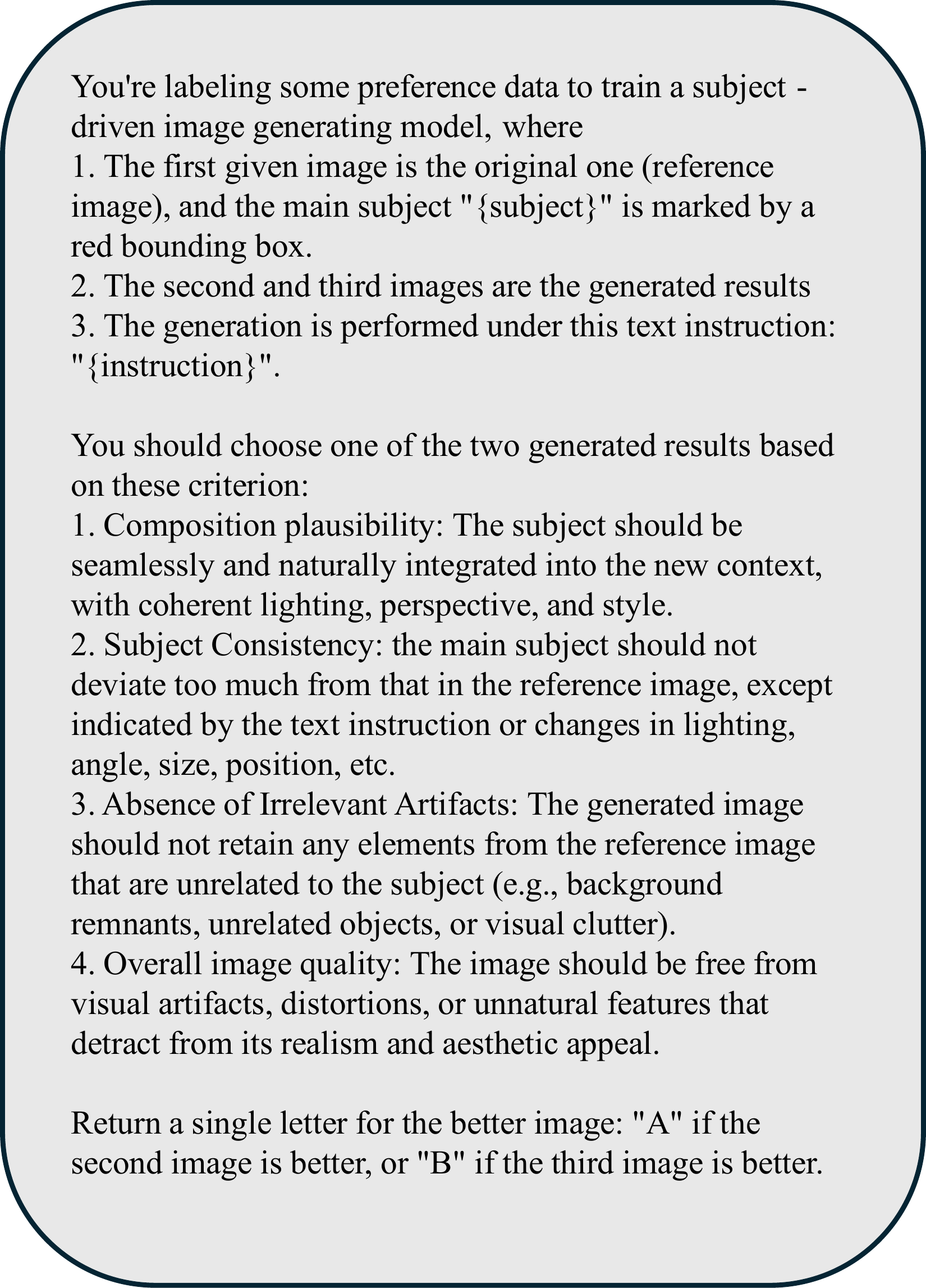}
    \caption{The prompt used to train the reward model.}
    \label{fig:prompt_rm}
\end{figure}

The prompts used to generate the negative editing instructions $\tilde{c_T}$ and to elicit pairwise judgements from $\phi$ are provided in \cref{fig:prompt_corrupt} and \cref{fig:prompt_rm}, respectively.

\section{Prompt for VLM in the TVD Module}
\begin{figure}[thpb]
    \centering
    \includegraphics[width=1\linewidth]{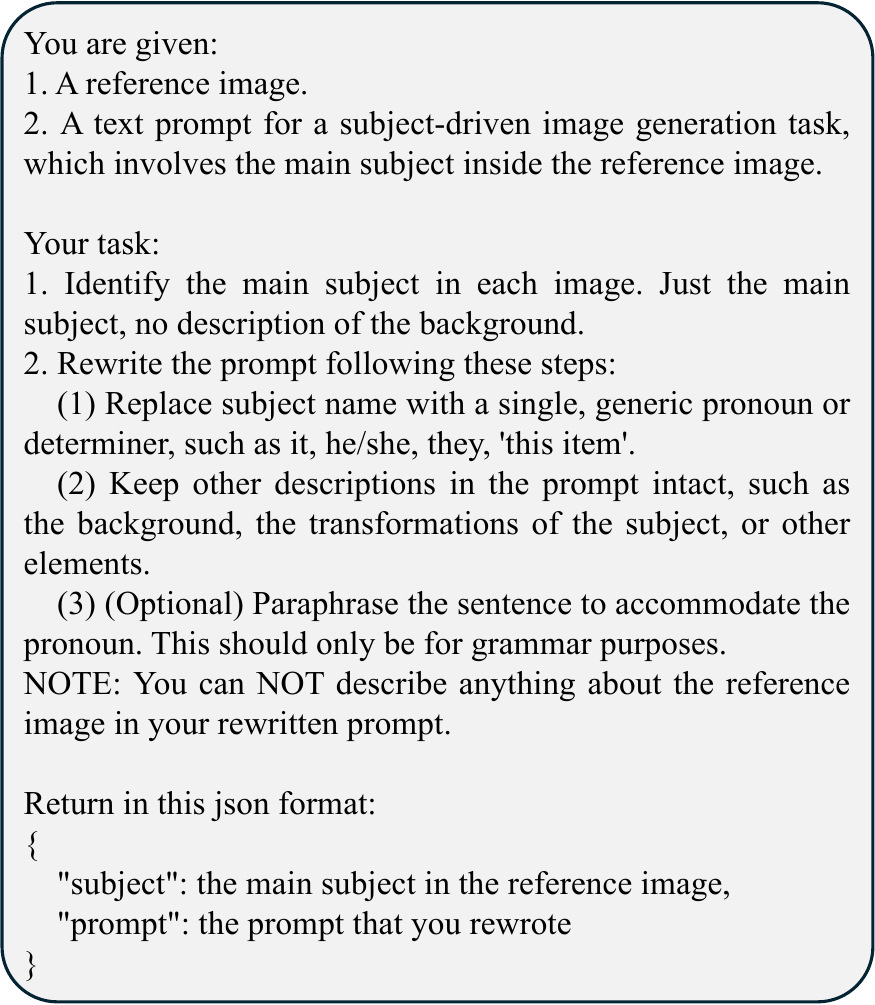}
    \caption{The prompt used in the TVD module.}
    \label{fig:prompt_tvd}
\end{figure}

We utilize Qwen2.5-VL-72B-Instruct as the VLM for textual-visual decoupling. The model is assigned two primary tasks: (1) identifying the main subject by jointly analyzing the reference image and the input text prompt, and (2) rewriting the prompt such that the identified subject is replaced with a generic pronoun. The specific prompt used to guide VLM in this process is provided in \cref{fig:prompt_tvd}.

% {
%     \small
%     \bibliographystyle{ieeenat_fullname}
%     \bibliography{suppl}
% }

%% file: main.bib
@String(CVPR= {IEEE Conf. Comput. Vis. Pattern Recog.})

@String(NIPS= {Adv. Neural Inform. Process. Syst.})

@String(CVPR  = {CVPR})

@String(NIPS  = {NeurIPS})

@inproceedings{wang2024ms,
  title={MS-Diffusion: Multi-subject Zero-shot Image Personalization with Layout Guidance},
  author={Wang, Xierui and Fu, Siming and Huang, Qihan and He, Wanggui and Jiang, Hao},
  booktitle={The Thirteenth International Conference on Learning Representations},
  year={2024}
}

@inproceedings{zhang2024ssr,
  title={Ssr-encoder: Encoding selective subject representation for subject-driven generation},
  author={Zhang, Yuxuan and Song, Yiren and Liu, Jiaming and Wang, Rui and Yu, Jinpeng and Tang, Hao and Li, Huaxia and Tang, Xu and Hu, Yao and Pan, Han and others},
  booktitle={Proceedings of the IEEE/CVF Conference on Computer Vision and Pattern Recognition},
  pages={8069--8078},
  year={2024}
}

@article{mao2025ace++,
  title={Ace++: Instruction-based image creation and editing via context-aware content filling},
  author={Mao, Chaojie and Zhang, Jingfeng and Pan, Yulin and Jiang, Zeyinzi and Han, Zhen and Liu, Yu and Zhou, Jingren},
  journal={arXiv preprint arXiv:2501.02487},
  year={2025}
}

@article{tan2024ominicontrol,
  title={Ominicontrol: Minimal and universal control for diffusion transformer},
  author={Tan, Zhenxiong and Liu, Songhua and Yang, Xingyi and Xue, Qiaochu and Wang, Xinchao},
  journal={arXiv preprint arXiv:2411.15098},
  year={2024}
}

@article{tan2025ominicontrol2,
  title={Ominicontrol2: Efficient conditioning for diffusion transformers},
  author={Tan, Zhenxiong and Xue, Qiaochu and Yang, Xingyi and Liu, Songhua and Wang, Xinchao},
  journal={arXiv preprint arXiv:2503.08280},
  year={2025}
}

@inproceedings{chen2025unireal,
  title={Unireal: Universal image generation and editing via learning real-world dynamics},
  author={Chen, Xi and Zhang, Zhifei and Zhang, He and Zhou, Yuqian and Kim, Soo Ye and Liu, Qing and Li, Yijun and Zhang, Jianming and Zhao, Nanxuan and Wang, Yilin and others},
  booktitle={Proceedings of the Computer Vision and Pattern Recognition Conference},
  pages={12501--12511},
  year={2025}
}

@inproceedings{xiao2025omnigen,
  title={Omnigen: Unified image generation},
  author={Xiao, Shitao and Wang, Yueze and Zhou, Junjie and Yuan, Huaying and Xing, Xingrun and Yan, Ruiran and Li, Chaofan and Wang, Shuting and Huang, Tiejun and Liu, Zheng},
  booktitle={Proceedings of the Computer Vision and Pattern Recognition Conference},
  pages={13294--13304},
  year={2025}
}

@article{wu2025less,
  title={Less-to-more generalization: Unlocking more controllability by in-context generation},
  author={Wu, Shaojin and Huang, Mengqi and Wu, Wenxu and Cheng, Yufeng and Ding, Fei and He, Qian},
  journal={arXiv preprint arXiv:2504.02160},
  year={2025}
}

@inproceedings{huang2024realcustom,
  title={RealCustom: narrowing real text word for real-time open-domain text-to-image customization},
  author={Huang, Mengqi and Mao, Zhendong and Liu, Mingcong and He, Qian and Zhang, Yongdong},
  booktitle={Proceedings of the IEEE/CVF Conference on Computer Vision and Pattern Recognition},
  pages={7476--7485},
  year={2024}
}

@article{mao2024realcustom++,
  title={Realcustom++: Representing images as real-word for real-time customization},
  author={Mao, Zhendong and Huang, Mengqi and Ding, Fei and Liu, Mingcong and He, Qian and Zhang, Yongdong},
  journal={arXiv preprint arXiv:2408.09744},
  year={2024}
}

@article{croitoru2023diffusion,
  title={Diffusion models in vision: A survey},
  author={Croitoru, Florinel-Alin and Hondru, Vlad and Ionescu, Radu Tudor and Shah, Mubarak},
  journal={IEEE Transactions on Pattern Analysis and Machine Intelligence},
  volume={45},
  number={9},
  pages={10850--10869},
  year={2023},
  publisher={IEEE}
}

@article{gal2022image,
  title={An image is worth one word: Personalizing text-to-image generation using textual inversion},
  author={Gal, Rinon and Alaluf, Yuval and Atzmon, Yuval and Patashnik, Or and Bermano, Amit H and Chechik, Gal and Cohen-Or, Daniel},
  journal={arXiv preprint arXiv:2208.01618},
  year={2022}
}

@inproceedings{ruiz2023dreambooth,
  title={Dreambooth: Fine tuning text-to-image diffusion models for subject-driven generation},
  author={Ruiz, Nataniel and Li, Yuanzhen and Jampani, Varun and Pritch, Yael and Rubinstein, Michael and Aberman, Kfir},
  booktitle={Proceedings of the IEEE/CVF conference on computer vision and pattern recognition},
  pages={22500--22510},
  year={2023}
}

@article{ye2023ip,
  title={Ip-adapter: Text compatible image prompt adapter for text-to-image diffusion models},
  author={Ye, Hu and Zhang, Jun and Liu, Sibo and Han, Xiao and Yang, Wei},
  journal={arXiv preprint arXiv:2308.06721},
  year={2023}
}

@inproceedings{wei2023elite,
  title={Elite: Encoding visual concepts into textual embeddings for customized text-to-image generation},
  author={Wei, Yuxiang and Zhang, Yabo and Ji, Zhilong and Bai, Jinfeng and Zhang, Lei and Zuo, Wangmeng},
  booktitle={Proceedings of the IEEE/CVF International Conference on Computer Vision},
  pages={15943--15953},
  year={2023}
}

@article{li2023blip,
  title={Blip-diffusion: Pre-trained subject representation for controllable text-to-image generation and editing},
  author={Li, Dongxu and Li, Junnan and Hoi, Steven},
  journal={Advances in Neural Information Processing Systems},
  volume={36},
  pages={30146--30166},
  year={2023}
}

@misc{mou2025dreamo,
      title={DreamO: A Unified Framework for Image Customization}, 
      author={Chong Mou and Yanze Wu and Wenxu Wu and Zinan Guo and Pengze Zhang and Yufeng Cheng and Yiming Luo and Fei Ding and Shiwen Zhang and Xinghui Li and Mengtian Li and Mingcong Liu and Yi Zhang and Shaojin Wu and Songtao Zhao and Jian Zhang and Qian He and Xinglong Wu},
      year={2025},
      eprint={2504.16915},
      archivePrefix={arXiv},
      primaryClass={cs.CV},
      url={https://arxiv.org/abs/2504.16915}, 
}

@misc{caron2021emerging,
      title={Emerging Properties in Self-Supervised Vision Transformers}, 
      author={Mathilde Caron and Hugo Touvron and Ishan Misra and Hervé Jégou and Julien Mairal and Piotr Bojanowski and Armand Joulin},
      year={2021},
      eprint={2104.14294},
      archivePrefix={arXiv},
      primaryClass={cs.CV},
      url={https://arxiv.org/abs/2104.14294}, 
}

@misc{kirillov2023segment,
      title={Segment Anything}, 
      author={Alexander Kirillov and Eric Mintun and Nikhila Ravi and Hanzi Mao and Chloe Rolland and Laura Gustafson and Tete Xiao and Spencer Whitehead and Alexander C. Berg and Wan-Yen Lo and Piotr Dollár and Ross Girshick},
      year={2023},
      eprint={2304.02643},
      archivePrefix={arXiv},
      primaryClass={cs.CV},
      url={https://arxiv.org/abs/2304.02643}, 
}

@misc{Emu2,
        title={Generative Multimodal Models are In-Context Learners}, 
        author={Quan Sun and Yufeng Cui and Xiaosong Zhang and Fan Zhang and Qiying Yu and Zhengxiong Luo and Yueze Wang and Yongming Rao and Jingjing Liu and Tiejun Huang and Xinlong Wang},
        publisher={arXiv preprint arXiv:2312.13286},
        year={2023}
  }

@article{xu2023imagereward,
  title={Imagereward: Learning and evaluating human preferences for text-to-image generation},
  author={Xu, Jiazheng and Liu, Xiao and Wu, Yuchen and Tong, Yuxuan and Li, Qinkai and Ding, Ming and Tang, Jie and Dong, Yuxiao},
  journal={Advances in Neural Information Processing Systems},
  volume={36},
  pages={15903--15935},
  year={2023}
}

@inproceedings{radford2021learning,
  title={Learning transferable visual models from natural language supervision},
  author={Radford, Alec and Kim, Jong Wook and Hallacy, Chris and Ramesh, Aditya and Goh, Gabriel and Agarwal, Sandhini and Sastry, Girish and Askell, Amanda and Mishkin, Pamela and Clark, Jack and others},
  booktitle={International conference on machine learning},
  pages={8748--8763},
  year={2021},
  organization={PmLR}
}

@inproceedings{wu2023human,
  title={Human preference score: Better aligning text-to-image models with human preference},
  author={Wu, Xiaoshi and Sun, Keqiang and Zhu, Feng and Zhao, Rui and Li, Hongsheng},
  booktitle={Proceedings of the IEEE/CVF International Conference on Computer Vision},
  pages={2096--2105},
  year={2023}
}

@inproceedings{rombach2022high,
  title={High-resolution image synthesis with latent diffusion models},
  author={Rombach, Robin and Blattmann, Andreas and Lorenz, Dominik and Esser, Patrick and Ommer, Bj{\"o}rn},
  booktitle={Proceedings of the IEEE/CVF conference on computer vision and pattern recognition},
  pages={10684--10695},
  year={2022}
}

@inproceedings{peebles2023scalable,
  title={Scalable diffusion models with transformers},
  author={Peebles, William and Xie, Saining},
  booktitle={Proceedings of the IEEE/CVF international conference on computer vision},
  pages={4195--4205},
  year={2023}
}

@misc{labs2025flux1kontextflowmatching,
      title={FLUX.1 Kontext: Flow Matching for In-Context Image Generation and Editing in Latent Space},
      author={Black Forest Labs and Stephen Batifol and Andreas Blattmann and Frederic Boesel and Saksham Consul and Cyril Diagne and Tim Dockhorn and Jack English and Zion English and Patrick Esser and Sumith Kulal and Kyle Lacey and Yam Levi and Cheng Li and Dominik Lorenz and Jonas Müller and Dustin Podell and Robin Rombach and Harry Saini and Axel Sauer and Luke Smith},
      year={2025},
      eprint={2506.15742},
      archivePrefix={arXiv},
      primaryClass={cs.GR},
      url={https://arxiv.org/abs/2506.15742},
}

@misc{flux2024,
    author={Black Forest Labs},
    title={FLUX},
    year={2024},
    howpublished={\url{https://github.com/black-forest-labs/flux}},
}

@inproceedings{liu2024grounding,
  title={Grounding dino: Marrying dino with grounded pre-training for open-set object detection},
  author={Liu, Shilong and Zeng, Zhaoyang and Ren, Tianhe and Li, Feng and Zhang, Hao and Yang, Jie and Jiang, Qing and Li, Chunyuan and Yang, Jianwei and Su, Hang and others},
  booktitle={European conference on computer vision},
  pages={38--55},
  year={2024},
  organization={Springer}
}

@article{guo2025deepseek,
  title={Deepseek-r1: Incentivizing reasoning capability in llms via reinforcement learning},
  author={Guo, Daya and Yang, Dejian and Zhang, Haowei and Song, Junxiao and Zhang, Ruoyu and Xu, Runxin and Zhu, Qihao and Ma, Shirong and Wang, Peiyi and Bi, Xiao and others},
  journal={arXiv preprint arXiv:2501.12948},
  year={2025}
}

@inproceedings{wallace2024diffusion,
  title={Diffusion model alignment using direct preference optimization},
  author={Wallace, Bram and Dang, Meihua and Rafailov, Rafael and Zhou, Linqi and Lou, Aaron and Purushwalkam, Senthil and Ermon, Stefano and Xiong, Caiming and Joty, Shafiq and Naik, Nikhil},
  booktitle={Proceedings of the IEEE/CVF Conference on Computer Vision and Pattern Recognition},
  pages={8228--8238},
  year={2024}
}

@article{xue2025dancegrpo,
  title={DanceGRPO: Unleashing GRPO on Visual Generation},
  author={Xue, Zeyue and Wu, Jie and Gao, Yu and Kong, Fangyuan and Zhu, Lingting and Chen, Mengzhao and Liu, Zhiheng and Liu, Wei and Guo, Qiushan and Huang, Weilin and others},
  journal={arXiv preprint arXiv:2505.07818},
  year={2025}
}

@article{loshchilov2017decoupled,
  title={Decoupled weight decay regularization},
  author={Loshchilov, Ilya and Hutter, Frank},
  journal={arXiv preprint arXiv:1711.05101},
  year={2017}
}

@article{achiam2023gpt,
  title={Gpt-4 technical report},
  author={Achiam, Josh and Adler, Steven and Agarwal, Sandhini and Ahmad, Lama and Akkaya, Ilge and Aleman, Florencia Leoni and Almeida, Diogo and Altenschmidt, Janko and Altman, Sam and Anadkat, Shyamal and others},
  journal={arXiv preprint arXiv:2303.08774},
  year={2023}
}

@article{zhang2025diffusion,
  title={Diffusion model as a noise-aware latent reward model for step-level preference optimization},
  author={Zhang, Tao and Da, Cheng and Ding, Kun and Yang, Huan and Jin, Kun and Li, Yan and Gao, Tingting and Zhang, Di and Xiang, Shiming and Pan, Chunhong},
  journal={arXiv preprint arXiv:2502.01051},
  year={2025}
}

@article{black2023training,
  title={Training diffusion models with reinforcement learning},
  author={Black, Kevin and Janner, Michael and Du, Yilun and Kostrikov, Ilya and Levine, Sergey},
  journal={arXiv preprint arXiv:2305.13301},
  year={2023}
}

@article{fan2023dpok,
  title={Dpok: Reinforcement learning for fine-tuning text-to-image diffusion models},
  author={Fan, Ying and Watkins, Olivia and Du, Yuqing and Liu, Hao and Ryu, Moonkyung and Boutilier, Craig and Abbeel, Pieter and Ghavamzadeh, Mohammad and Lee, Kangwook and Lee, Kimin},
  journal={Advances in Neural Information Processing Systems},
  volume={36},
  pages={79858--79885},
  year={2023}
}

@article{shao2024deepseekmath,
  title={Deepseekmath: Pushing the limits of mathematical reasoning in open language models},
  author={Shao, Zhihong and Wang, Peiyi and Zhu, Qihao and Xu, Runxin and Song, Junxiao and Bi, Xiao and Zhang, Haowei and Zhang, Mingchuan and Li, YK and Wu, Yang and others},
  journal={arXiv preprint arXiv:2402.03300},
  year={2024}
}

@article{schulman2017proximal,
  title={Proximal policy optimization algorithms},
  author={Schulman, John and Wolski, Filip and Dhariwal, Prafulla and Radford, Alec and Klimov, Oleg},
  journal={arXiv preprint arXiv:1707.06347},
  year={2017}
}

@article{liu2025flow,
  title={Flow-grpo: Training flow matching models via online rl},
  author={Liu, Jie and Liu, Gongye and Liang, Jiajun and Li, Yangguang and Liu, Jiaheng and Wang, Xintao and Wan, Pengfei and Zhang, Di and Ouyang, Wanli},
  journal={arXiv preprint arXiv:2505.05470},
  year={2025}
}

@article{su2024roformer,
  title={Roformer: Enhanced transformer with rotary position embedding},
  author={Su, Jianlin and Ahmed, Murtadha and Lu, Yu and Pan, Shengfeng and Bo, Wen and Liu, Yunfeng},
  journal={Neurocomputing},
  volume={568},
  pages={127063},
  year={2024},
  publisher={Elsevier}
}

@misc{bai2025qwen25vl,
      title={Qwen2.5-VL Technical Report}, 
      author={Shuai Bai and Keqin Chen and Xuejing Liu and Jialin Wang and Wenbin Ge and Sibo Song and Kai Dang and Peng Wang and Shijie Wang and Jun Tang and Humen Zhong and Yuanzhi Zhu and Mingkun Yang and Zhaohai Li and Jianqiang Wan and Pengfei Wang and Wei Ding and Zheren Fu and Yiheng Xu and Jiabo Ye and Xi Zhang and Tianbao Xie and Zesen Cheng and Hang Zhang and Zhibo Yang and Haiyang Xu and Junyang Lin},
      year={2025},
      eprint={2502.13923},
      archivePrefix={arXiv},
      primaryClass={cs.CV},
      url={https://arxiv.org/abs/2502.13923}, 
}

@inproceedings{esser2024scaling,
  title={Scaling rectified flow transformers for high-resolution image synthesis},
  author={Esser, Patrick and Kulal, Sumith and Blattmann, Andreas and Entezari, Rahim and M{\"u}ller, Jonas and Saini, Harry and Levi, Yam and Lorenz, Dominik and Sauer, Axel and Boesel, Frederic and others},
  booktitle={Forty-first international conference on machine learning},
  year={2024}
}

@INPROCEEDINGS{shin2025diptych,
  author={Shin, Chaehun and Choi, Jooyoung and Kim, Heeseung and Yoon, Sungroh},
  booktitle={2025 IEEE/CVF Conference on Computer Vision and Pattern Recognition (CVPR)}, 
  title={Large-Scale Text-to-Image Model with Inpainting is a Zero-Shot Subject-Driven Image Generator}, 
  year={2025},
  volume={},
  number={},
  pages={7986-7996},
  doi={10.1109/CVPR52734.2025.00748}
  }

@inproceedings{miao2024rpo,
author = {Miao, Yanting and Loh, William and Kothawade, Suraj and Poupart, Pascal and Rashwan, Abdullah and Li, Yeqing},
title = {Subject-driven text-to-image generation via preference-based reinforcement learning},
year = {2024},
isbn = {9798331314385},
publisher = {Curran Associates Inc.},
address = {Red Hook, NY, USA},
booktitle = {Proceedings of the 38th International Conference on Neural Information Processing Systems},
articleno = {3928},
numpages = {29},
location = {Vancouver, BC, Canada},
series = {NIPS '24}
}

@article{xiao2024fastcomposer,
author = {Xiao, Guangxuan and Yin, Tianwei and Freeman, William T. and Durand, Fr\'{e}do and Han, Song},
title = {FastComposer: Tuning-Free Multi-subject Image Generation with Localized Attention},
year = {2024},
issue_date = {Mar 2025},
publisher = {Kluwer Academic Publishers},
address = {USA},
volume = {133},
number = {3},
issn = {0920-5691},
url = {https://doi.org/10.1007/s11263-024-02227-z},
doi = {10.1007/s11263-024-02227-z},
journal = {Int. J. Comput. Vision},
month = sep,
pages = {1175–1194},
numpages = {20}
}

@INPROCEEDINGS {li2024photomaker,
    author = { Li, Zhen and Cao, Mingdeng and Wang, Xintao and Qi, Zhongang and Cheng, Ming-Ming and Shan, Ying },
    booktitle = { 2024 IEEE/CVF Conference on Computer Vision and Pattern Recognition (CVPR) },
    title = {{ PhotoMaker: Customizing Realistic Human Photos via Stacked ID Embedding }},
    year = {2024},
    volume = {},
    ISSN = {},
    pages = {8640-8650},
    doi = {10.1109/CVPR52733.2024.00825},
    url = {https://doi.ieeecomputersociety.org/10.1109/CVPR52733.2024.00825},
    publisher = {IEEE Computer Society},
    address = {Los Alamitos, CA, USA},
    month =Jun
}

@inproceedings{wang2024moa,
    author = {Wang, Kuan-Chieh and Ostashev, Daniil and Fang, Yuwei and Tulyakov, Sergey and Aberman, Kfir},
    title = {MoA: Mixture-of-Attention for Subject-Context Disentanglement in Personalized Image Generation},
    year = {2024},
    isbn = {9798400711312},
    publisher = {Association for Computing Machinery},
    address = {New York, NY, USA},
    url = {https://doi.org/10.1145/3680528.3687662},
    doi = {10.1145/3680528.3687662},
    booktitle = {SIGGRAPH Asia 2024 Conference Papers},
    articleno = {3},
    numpages = {12},
    location = {Tokyo, Japan},
    series = {SA '24}
}

@inproceedings{patashnik2025nestedattention,
author = {Patashnik, Or and Gal, Rinon and Ostashev, Daniil and Tulyakov, Sergey and Aberman, Kfir and Cohen-Or, Daniel},
title = {Nested Attention: Semantic-aware Attention Values for Concept Personalization},
year = {2025},
isbn = {9798400715402},
publisher = {Association for Computing Machinery},
address = {New York, NY, USA},
url = {https://doi.org/10.1145/3721238.3730634},
doi = {10.1145/3721238.3730634},
booktitle = {Proceedings of the Special Interest Group on Computer Graphics and Interactive Techniques Conference Conference Papers},
articleno = {6},
numpages = {12},
location = {
},
series = {SIGGRAPH Conference Papers '25}
}

@misc{bai2025qwen3vl,
      title={Qwen3-VL Technical Report}, 
      author={Shuai Bai and Yuxuan Cai and Ruizhe Chen and Keqin Chen and Xionghui Chen and Zesen Cheng and Lianghao Deng and Wei Ding and Chang Gao and Chunjiang Ge and Wenbin Ge and Zhifang Guo and Qidong Huang and Jie Huang and Fei Huang and Binyuan Hui and Shutong Jiang and Zhaohai Li and Mingsheng Li and Mei Li and Kaixin Li and Zicheng Lin and Junyang Lin and Xuejing Liu and Jiawei Liu and Chenglong Liu and Yang Liu and Dayiheng Liu and Shixuan Liu and Dunjie Lu and Ruilin Luo and Chenxu Lv and Rui Men and Lingchen Meng and Xuancheng Ren and Xingzhang Ren and Sibo Song and Yuchong Sun and Jun Tang and Jianhong Tu and Jianqiang Wan and Peng Wang and Pengfei Wang and Qiuyue Wang and Yuxuan Wang and Tianbao Xie and Yiheng Xu and Haiyang Xu and Jin Xu and Zhibo Yang and Mingkun Yang and Jianxin Yang and An Yang and Bowen Yu and Fei Zhang and Hang Zhang and Xi Zhang and Bo Zheng and Humen Zhong and Jingren Zhou and Fan Zhou and Jing Zhou and Yuanzhi Zhu and Ke Zhu},
      year={2025},
      eprint={2511.21631},
      archivePrefix={arXiv},
      primaryClass={cs.CV},
      url={https://arxiv.org/abs/2511.21631}, 
}

@misc{wu2025qwenimage,
      title={Qwen-Image Technical Report}, 
      author={Chenfei Wu and Jiahao Li and Jingren Zhou and Junyang Lin and Kaiyuan Gao and Kun Yan and Sheng-ming Yin and Shuai Bai and Xiao Xu and Yilei Chen and Yuxiang Chen and Zecheng Tang and Zekai Zhang and Zhengyi Wang and An Yang and Bowen Yu and Chen Cheng and Dayiheng Liu and Deqing Li and Hang Zhang and Hao Meng and Hu Wei and Jingyuan Ni and Kai Chen and Kuan Cao and Liang Peng and Lin Qu and Minggang Wu and Peng Wang and Shuting Yu and Tingkun Wen and Wensen Feng and Xiaoxiao Xu and Yi Wang and Yichang Zhang and Yongqiang Zhu and Yujia Wu and Yuxuan Cai and Zenan Liu},
      year={2025},
      eprint={2508.02324},
      archivePrefix={arXiv},
      primaryClass={cs.CV},
      url={https://arxiv.org/abs/2508.02324}, 
}

@misc{seedream2025seedream4,
      title={Seedream 4.0: Toward Next-generation Multimodal Image Generation}, 
      author={Team Seedream and : and Yunpeng Chen and Yu Gao and Lixue Gong and Meng Guo and Qiushan Guo and Zhiyao Guo and Xiaoxia Hou and Weilin Huang and Yixuan Huang and Xiaowen Jian and Huafeng Kuang and Zhichao Lai and Fanshi Li and Liang Li and Xiaochen Lian and Chao Liao and Liyang Liu and Wei Liu and Yanzuo Lu and Zhengxiong Luo and Tongtong Ou and Guang Shi and Yichun Shi and Shiqi Sun and Yu Tian and Zhi Tian and Peng Wang and Rui Wang and Xun Wang and Ye Wang and Guofeng Wu and Jie Wu and Wenxu Wu and Yonghui Wu and Xin Xia and Xuefeng Xiao and Shuang Xu and Xin Yan and Ceyuan Yang and Jianchao Yang and Zhonghua Zhai and Chenlin Zhang and Heng Zhang and Qi Zhang and Xinyu Zhang and Yuwei Zhang and Shijia Zhao and Wenliang Zhao and Wenjia Zhu},
      year={2025},
      eprint={2509.20427},
      archivePrefix={arXiv},
      primaryClass={cs.CV},
      url={https://arxiv.org/abs/2509.20427}, 
}
